\documentclass{article}

\usepackage{microtype}
\usepackage{graphicx}
\usepackage{subcaption}
\usepackage{booktabs} 

\usepackage{hyperref}
\usepackage{makecell}

\usepackage[accepted]{icml2026}

\usepackage{amsmath}
\usepackage{amssymb}
\usepackage{mathtools}
\usepackage{amsthm}

\usepackage{booktabs}
\usepackage{multirow}
\usepackage{graphicx} 

\usepackage[capitalize,noabbrev]{cleveref}

\theoremstyle{plain}

\theoremstyle{definition}

\theoremstyle{remark}

\usepackage[textsize=tiny]{todonotes}

\begin{document}

\twocolumn[
  \icmltitle{ACFormer: Mitigating Non-linearity with Auto Convolutional Encoder \\
 for Time Series Forecasting }



  \icmlsetsymbol{equal}{*}

  \begin{icmlauthorlist}
    \icmlauthor{Gawon Lee}{pusan}
    \icmlauthor{Hanbyeol Park}{pusan}
    \icmlauthor{Minseop Kim}{pusan}
    \icmlauthor{Dohee Kim}{changwon}
    \icmlauthor{Hyerim Bae}{pusan}
  \end{icmlauthorlist}

  \icmlaffiliation{pusan}{Pusan National University, Busan, Republic of Korea}
  \icmlaffiliation{changwon}{Pusan National University, Changwon, Republic of Korea}

  \icmlcorrespondingauthor{Gawon Lee}{monago@pusan.ac.kr}
  \icmlcorrespondingauthor{Hanbyeol Park}{pb104@pusan.ac.kr}
  \icmlcorrespondingauthor{Minseop Kim}{rlaals7349@pusan.ac.kr}
  \icmlcorrespondingauthor{Dohee Kim}{kimdohee@changwon.ac.kr}
  \icmlcorrespondingauthor{Hyerim Bae}{hrbae@pusan.ac.kr}

  \icmlkeywords{Machine Learning, ICML}

  \vskip 0.3in
]



\printAffiliationsAndNotice{}  

\begin{abstract}
Time series forecasting (TSF) faces challenges in modeling complex intra-channel temporal dependencies and inter-channel correlations. Although recent research has highlighted the efficiency of linear architectures in capturing global trends, these models often struggle with non-linear signals. To address this gap,  we conducted a systematic receptive field analysis of convolutional neural network (CNN) TSF models. We introduce the "individual receptive field" to uncover granular structural dependencies, revealing that convolutional layers act as feature extractors that mirror channel-wise attention while exhibiting superior robustness to non-linear fluctuations. Based on these insights, we propose ACFormer, an architecture designed to reconcile the efficiency of linear projections with the non-linear feature-extraction power of convolutions. ACFormer captures fine-grained information through a shared compression module, preserves temporal locality via gated attention, and reconstructs variable-specific temporal patterns using an independent patch expansion layer. Extensive experiments on multiple benchmark datasets demonstrate that ACFormer consistently achieves state-of-the-art performance, effectively mitigating the inherent drawbacks of linear models in capturing high-frequency components.
\end{abstract}

\section{Introduction}

\begin{figure}
  \centering
  \includegraphics[width=1.0\linewidth]{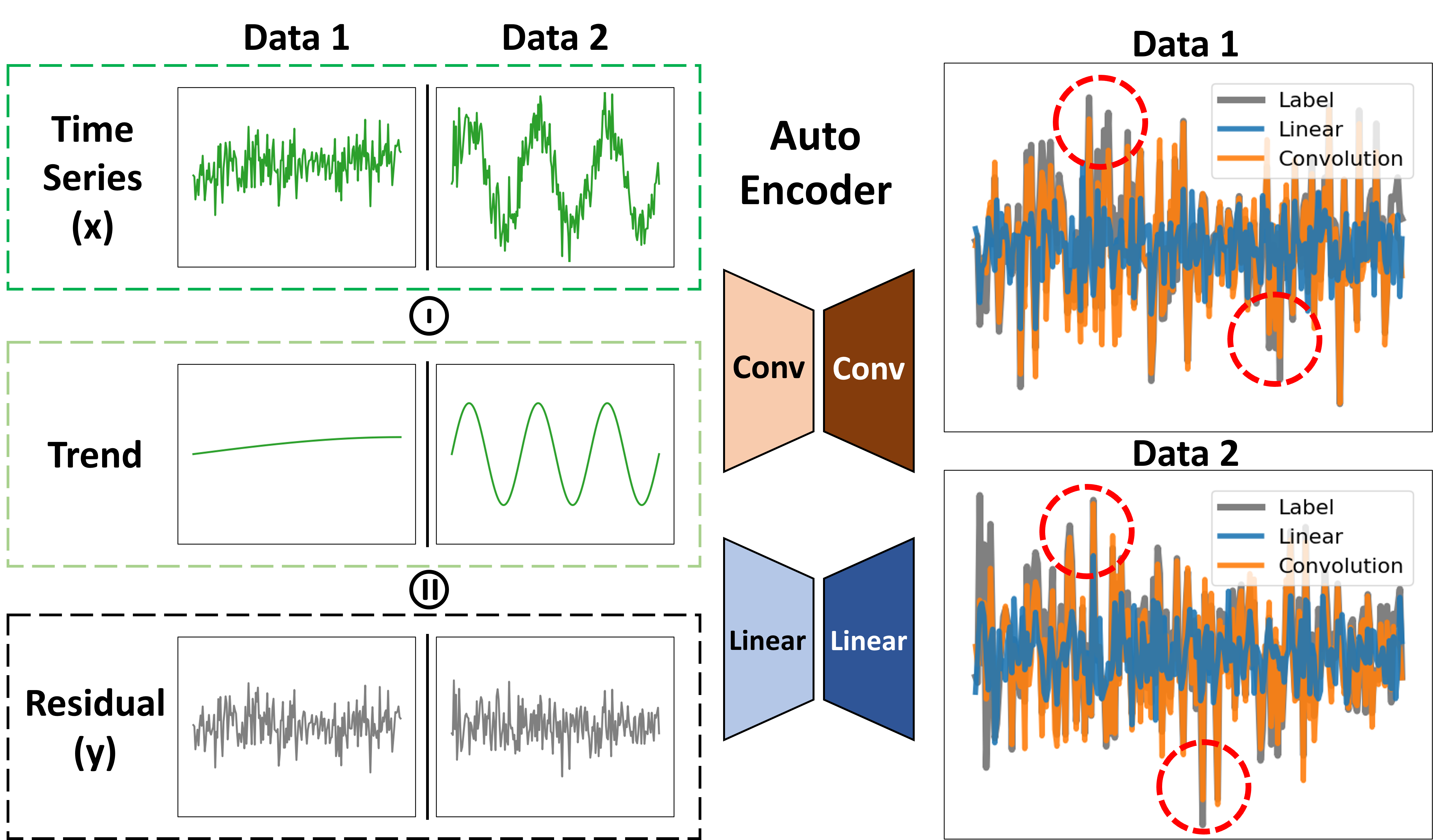}
  \caption{Residual extraction performance comparison between convolutional (Orange) and linear layers (Blue). The linear layers fail to reconstruct high-frequency components (marked by red circles), whereas convolutional layers succeed.}
  \label{fig:AutoEncExp}
\end{figure}

Time series forecasting (TSF) is essential for decision-making across diverse industrial domains, including meteorology\cite{Weather}, financial engineering\cite{Finance}, and resource management \cite{Sensor}. The fundamental challenge in TSF lies in accurately modeling both intra-channel temporal dependencies and inter-channel correlations. This complexity is further compounded in multivariate settings, in which temporal resolutions and influences vary across channels. Specifically, cross-channel influences often manifest across disparate time scales: a fluctuation in one variable may exert an immediate impact on another, whereas its influence on a separate channel may only emerge after a significant lag. Capturing these heterogeneous lag patterns remains a significant hurdle for robust predictive modeling.

Following by the success of the Transformer architecture in natural language processing, recent research has sought to leverage self-attention mechanisms to model complex temporal dependencies. Early adaptations typically relied on token embedding methods. However, these models often discretize continuous time series into point-wise tokens. This tokenization approach can limit the efficacy of attention mechanisms by restricting their receptive fields to individual time steps \cite{Autoformer}, thereby neglecting multi-resolution contextual information. Although convolutional layers have been integrated as embedding modules to capture local patterns, they often incur in significant computational overhead when the kernel sizes are expanded to encompass long-term dependencies.

To address these limitations, a recent paradigm shift has favored linear-based TSF architectures. These models demonstrate that simple linear layers can effectively capture the global periodicity and long-term trends with minimal complexity \cite{DLinear}. Consequently, many modern frameworks utilize linear encoders integrated with patch- or channel-wise attention, thereby replacing the traditional decoder with a lightweight projection layer for extrapolation \cite{PatchTST, iTransformer}.

Despite their high efficiency, linear models have two critical limitations. First, projecting sequences into high-dimensional spaces often disregards intrinsic sequential properties such as trends or residuals. Given the success of downsampling in time series analysis \cite{TimeMixerPP}, exploiting these inherent temporal properties is often more effective than high-dimensional expansion. Second, linear layers struggle with non-stationary data and high-frequency components. Although decomposition-based methods attempt to mitigate this, they frequently fail to capture the non-linearity of seasonal fluctuations, especially in multivariate settings in which disparate periodicities across channels can degrade predictive accuracy \cite{LinearMapping, Amplifier}.

In this paper, we bridge this gap by revisiting CNNs in TSF domain and analyzing the receptive fields of CNN-based TSF models. We demonstrate that the convolutional layers mirror channel-wise attention while providing superior robustness to non-linear data. Building upon these insights, we propose ACFormer, an architecture that utilizes auto-convolution layers to extract fine-grained features while maintaining the efficiency of linear projections through channel-wise attention.

The main contributions of this study can be summarized as follows:

\begin{itemize}
    \item We provide a comprehensive investigation into convolutional layers in TSF, demonstrating through receptive field analysis that they act as robust, non-linear alternatives to standard channel-wise attention.
    \item We introduce ACFormer, a novel architecture that reconciles linear projection efficiency with convolutional feature-extraction power using an auto-convolution mechanism.
    \item Extensive experimental results on real-world benchmarks demonstrate that ACFormer consistently outperforms existing baselines to achieve state-of-the-art(SOTA) performance.
\end{itemize}  

\section{Related work}

TSF has transitioned from modeling global dependencies to leveraging simplified linear structures and local semantic information. Early Transformer adaptations , such as Informer, Autoformer, and FEDformer, sought to capture long-range dependencies by addressing the quadratic complexity of self-attention through sparse attention, auto-correlation, or frequency-domain analysis \cite{Autoformer, Informer, FEDFormer}. However, these architectures often rely on point-wise tokenization, which fails to capture the local context and leaves models susceptible to non-stationary data \cite{Nonstationary}.

A significant paradigm shift emerged with DLinear, which demonstrated that simple linear layers could outperform complex Transformers by effectively modeling trends and seasonality \cite{DLinear}. This discovery led to current SOTA architectures like PatchTST and iTransformer, which utilize linear layers as fundamental modules for embedding and projections. Patch-wise embedding addresses point-wise limitations through patch-wise embedding to aggregate adjacent time steps \cite{DeformableTST, PatchTST, xPatch}, while sequence-wise embedding inverts the traditional framework to model inter-variable correlations via channel-wise attention \cite{iTransformer, SAMFormer}. Despite their success, linear embeddings often struggle with high-frequency, non-linear signal time series data.

To address these limitations, CNNs have been developed by leveraging the inductive bias of locality. While temporal convolutional networks utilize dilated convolutions to expand receptive fields, they often lose critical local information \cite{TCN}. Subsequent models, such as SCINet and TimesNet, introduced recursive downsampling and 2D-variation modeling to capture multi-resolution features \cite{SCINet, TimesNet}. Notably, ModernTCN argues that Transformer success stems from a large effective receptive field (ERF) rather than the attention mechanism itself. By utilizing large-kernel depth-wise convolutions, ModernTCN achieves an expansive ERF that rivals the Transformer while maintaining the feature extraction abilities of pure convolutional structure.

ACFormer builds on this receptive fields and expresivness focus of convolutional layers. While ModernTCN optimizes the ERF through large kernels, we propose integrating channel-wise attention for a more efficient receptive field expansion. Furthermore, we adopt an auto-convolutional encoder for non-linear feature extraction and amplification. This approach reconciles the high-efficiency standards of linear projections with the robust feature-extraction power characteristics of CNN layers.


\section{In depth CNN analysis}

\subsection{Preliminaries}
In TSF, input sequence $X\in\mathbb{R}^{\mathbf{S}\times\mathbf{C}}$ is provided to deep learning model and outputs $\hat{Y}\in\mathbb{R}^{\mathbf{P}\times \mathbf{C}}$, where $S$ is the input sequence length, $P$ is the predicted sequence length and $C$ is the channel size. The aim is to minimize the distance between predictions $\hat{Y}$ and true labels $Y\in\mathbb{R}^{\mathbf{P}\times \mathbf{C}}$.

\subsection{Individual Receptive Fields} 
\label{section:Individual_receptive_fields}

Conventional receptive field analysis for TSF averages gradients across all channels \cite{Effective_Receptive_Field}. The computation of conventional receptive fields can be expressed as:
\begin{equation}
    F=\frac{1}{C}\sum_{c=1}^{C}{\hat{y}_{P/2,c}}    
\end{equation}
where $\hat{y}_{P/2,c}$ is at the middle of the output sequence for channel $c$. Feature value $F$ represents the mean of the central points across all channels. The final receptive field $G$ is then obtained by dividing along the feature dimensions again and calculating the average, as follows:

\begin{equation}
    G=\frac{1}{C}\sum_{c=1}^C{\frac{\partial{F}}{\partial{x_{c}}}}
\end{equation}
where $G\in\mathbb{R}^{S}$ denotes the gradients of the feature map relative to $x$. However, the limitations of this conventional analysis are pronounced in multivariate time series data, in which averaging disregards critical inter-channel correlations. Unlike the three-channel RGB data typical in computer vision, multivariate time series comprise diverse, high-dimensional variables that interact through complex dependencies. By collapsing these dimensions, traditional methods fail to identify specific structural relationships and reference patterns between channels. 

To address this, we propose the "individual receptive field," a method that preserves the gradient values of each channel independently. By maintaining the integrity of the channel dimension rather than averaging it, this approach uncovers the true structural dependencies and provides a granular visualization of how specific input variables influence the output sequence. Computation of the individual receptive fields is as follows:
\begin{equation}
    IG = \Bigl\{\frac{\partial{\hat{y}_{P/2,c}}}{\partial{x}}\Bigl|c\in[1, 2, \dots, C] \Bigl\}
\end{equation}
where $IG\in\mathbb{R}^{S\times C\times C}$ preserving input and output channel dimensions. Unlike existing methods that rely on ReLU functions to isolate positive gradients, our approach preserves negative values to capture the full spectrum of input influences by utilizing min-max scaling for cross-channel normalization. 

Using ModernTCN as a representative convolutional network for individual receptive fields analysis, we identifies a subset of "pivot channels," highly selective variables whose dynamic influence drives predictions across the entire feature set. A visualization of individual receptive fields is presented in Appendix \ref{appendix:individual_receptive_field}. To quantify this relationship, we introduce "variance attention," a metric derived from the temporal variance of gradients:
\begin{equation}
    VA = \{Var(IG_{c,c'})|c,c'\in[1, 2, \dots, C]\}
\end{equation}
where $IG_{c,c'}\in\mathbb{R}^{S}$ is the sequence of gradient values for input channel $c$ with respect to the output sequence of channel $c'$. 

\begin{figure}%
    \centering
    \subfloat[\centering channel-wise attention]{{\includegraphics[width=.45\linewidth]{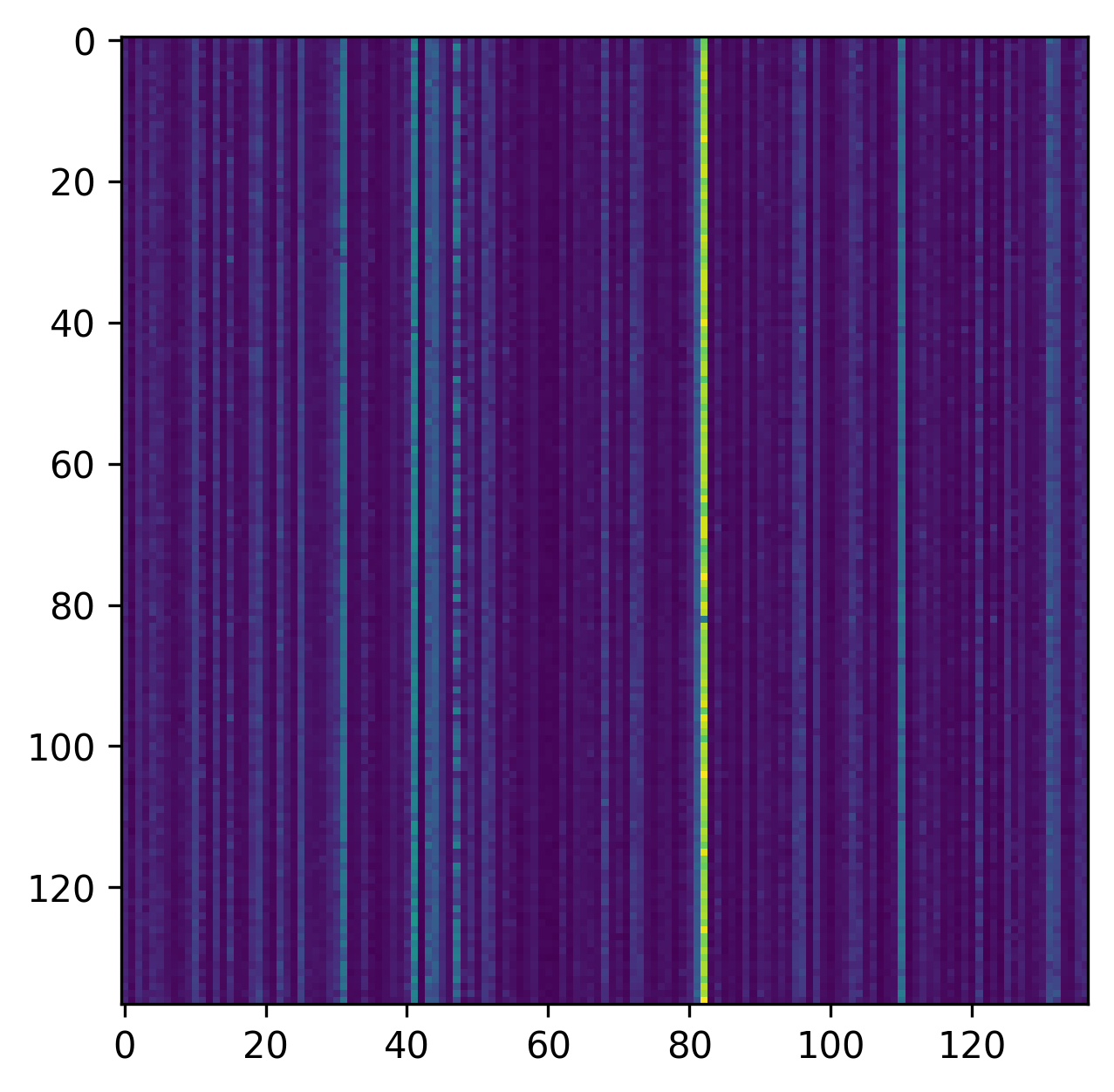} }}%
    \subfloat[\centering variance attention]{{\includegraphics[width=.45\linewidth]{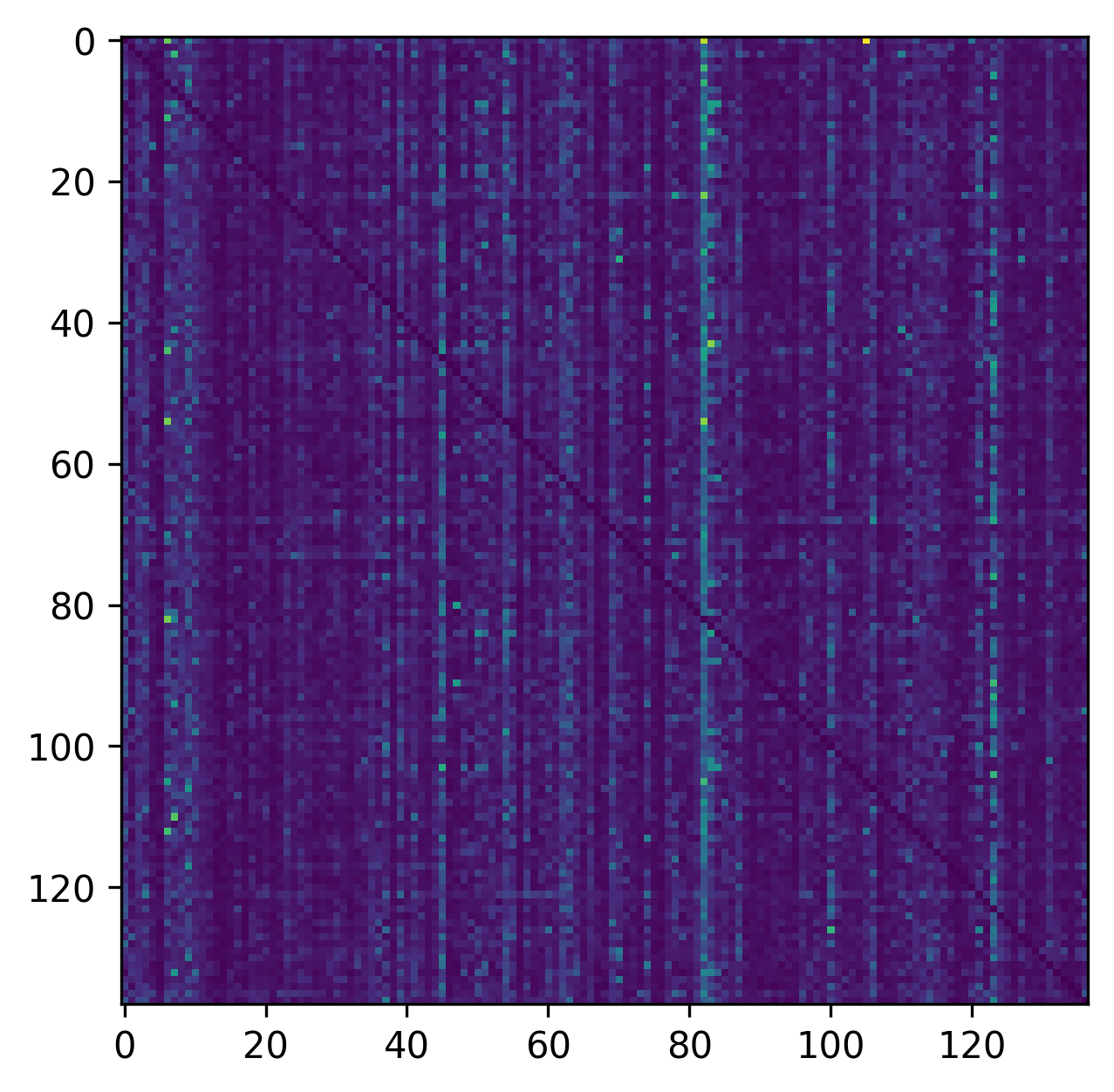} }}%
    \caption{Comparison between the (a) channel-wise attention from iTransformer and (b) our proposed variance attention from ModernTCN.}%
    \label{fig:Comp_CW_and_V}%
\end{figure}

Unlike static mean values, a high temporal variance effectively captures the dynamic sensitivity and predictive importance of a channel. By mapping this variance onto a channel-by-channel grid, we can directly compare the implicit structural learning of CNNs with the explicit channel-wise attention mechanisms found in architectures such as the iTransformer. Experimental results using the Solar Energy dataset (137 variables), as represented in Figure \ref{fig:Comp_CW_and_V} reveal a striking similarity between channel-wise attention and the proposed variance attention, as both mechanisms prioritize identical pivot channels.

This suggests that channel-wise attention functions as an explicit global modeling mechanism to expand the receptive field by identifying these pivot channels. While convolutional models must implicitly learn these dependencies through deeper architectures and temporal kernels, channel-wise attention isolates pivotal information more directly and with a significantly lower computational overhead. In summary, pivot channels drive predictive performance across both paradigms. However, explicit mapping via channel-wise attention offers a more targeted and efficient path for leveraging the pivoting information essential to multivariate forecasting.

\subsection{Non-Linearity of Convolutional Layers}  
\label{section:Non-linearity}
Although channel-wise attention efficiently identifies cross-channel dependencies, its ability to capture the intricate, non-linear dynamics of complex time series requires further investigation. Although channel-wise attention excel in relational modeling, they essentially perform weighted sums along channel dimensions. 

Such approach limits its capacity to address to specific, non-linear fluctuations at localized time steps, often resulting in high-frequency information loss. By contrast, convolutional layers possess a unique inductive bias for local temporal dependencies, allowing them to function as robust non-linear feature extractors.

To investigate these limitations, we conducted a comparative analysis between the linear and convolutional layers and evaluated their performance in isolating high-frequency components from structured data. The experimental process is illustrated in Figure \ref{fig:AutoEncExp}. We utilize synthetic data characterized by diverse periodicity and stochastic noise, formulated as:
\begin{equation}
    s_n = \{\mathbf{a}sin(2\mathbf{b}\pi i/l)+\epsilon_{n,i} | i\in[0,1,\dots,l-1]\}
  \label{eq:synthetic_dataset}
\end{equation}
where $\mathbf{a}$ represents random amplitude, $\mathbf{b}$ denotes random periodicity, $l$ is the sequence length, and $\epsilon \sim N(\mu=0,\sigma=1)$ represents Gaussian noise. A more detailed explanation of the experimental settings is provided in Appendix \ref{appendix:synthetic} The objective of the model is to map input $s$, containing both global trends and local residuals, to the isolated high-frequency residual component $\epsilon$. 

Our framework employs an auto-encoder architecture that compresses the input into a lower-dimensional latent space before projecting it back to its original dimensions. To isolate the effects of the underlying operations, we varied the architectural composition of the compression (encoder) and expansion (decoder) compartments using either linear or convolutional layers. This yielded four experimental configurations and each model was trained to minimize the mean squared error (MSE) between the predicted $\hat{\epsilon}$ and the true residuals $\epsilon$. As summarized in Table \ref{tab:Res_extraction}, the empirical results demonstrate a significant degradation in accuracy when linear compartments were utilized.

\begin{table}[t]
  \caption{Residual extraction accuracies for different compartments.}
  \label{tab:Res_extraction}
  \begin{center}
    \begin{small}
      \begin{sc}
        \begin{tabular}{ccccr}
          \toprule
          Enc   & Dec     & MAE       & MSE   \\
          \midrule
          Lin       & Lin       & 0.7567    & 0.6949 \\
          Lin       & Conv      & 0.7840    & 0.7043 \\
          Conv      & Lin       & 0.7878    & 0.7060 \\
          Conv      & Conv      & \textbf{0.0974}    & \textbf{0.2306} \\
          \bottomrule
        \end{tabular}
      \end{sc}
    \end{small}
  \end{center}
  \vskip -0.1in
\end{table}

The prediction results shown in Figure \ref{fig:AutoEncExp} further illustrates high-frequency information loss in linear models. The red circles in the plot indicate sudden spikes or residual values, which convolution models successfully predicts but the linear model fails. This suggests a fundamental limitation of linear layers in mapping the transformations required to separate high-frequency signals from the underlying trends.

To verify whether this bottleneck is inherent to the linear operation itself, we conducted an ablation study by inserting a linear projection layer between the convolutional encoder and decoder. Interestingly, as presented in Table \ref{tab:Res_extraction2}, the inclusion of this middle linear layer actually improved the performance compared with the all-convolutional baseline. This observation suggests that while hidden state representations can be manipulated linearly, the initial extraction of features from raw, non-stationary time series requires the non-linear inductive bias of convolutions. These findings justify the design of auto-convolution design, which leverages these non-linear properties while integrating channel-wise attention to model the relation between channels.

\begin{table}[t]
  \caption{Impact of inserting a linear projection layer between the convolutional encoder and decoder.}
  \label{tab:Res_extraction2}
  \begin{center}
    \begin{small}
      \begin{sc}
        \begin{tabular}{cccccr}
          \toprule
          Enc   & Proj    & Dec     & MAE       & MSE   \\
          \midrule
          Conv   & Lin        & Conv   & 0.0530    & 0.2061 \\
          \bottomrule
        \end{tabular}
      \end{sc}
    \end{small}
  \end{center}
  \vskip -0.1in
\end{table}


\begin{figure*}
    \centering
    \includegraphics[width=0.95\linewidth]{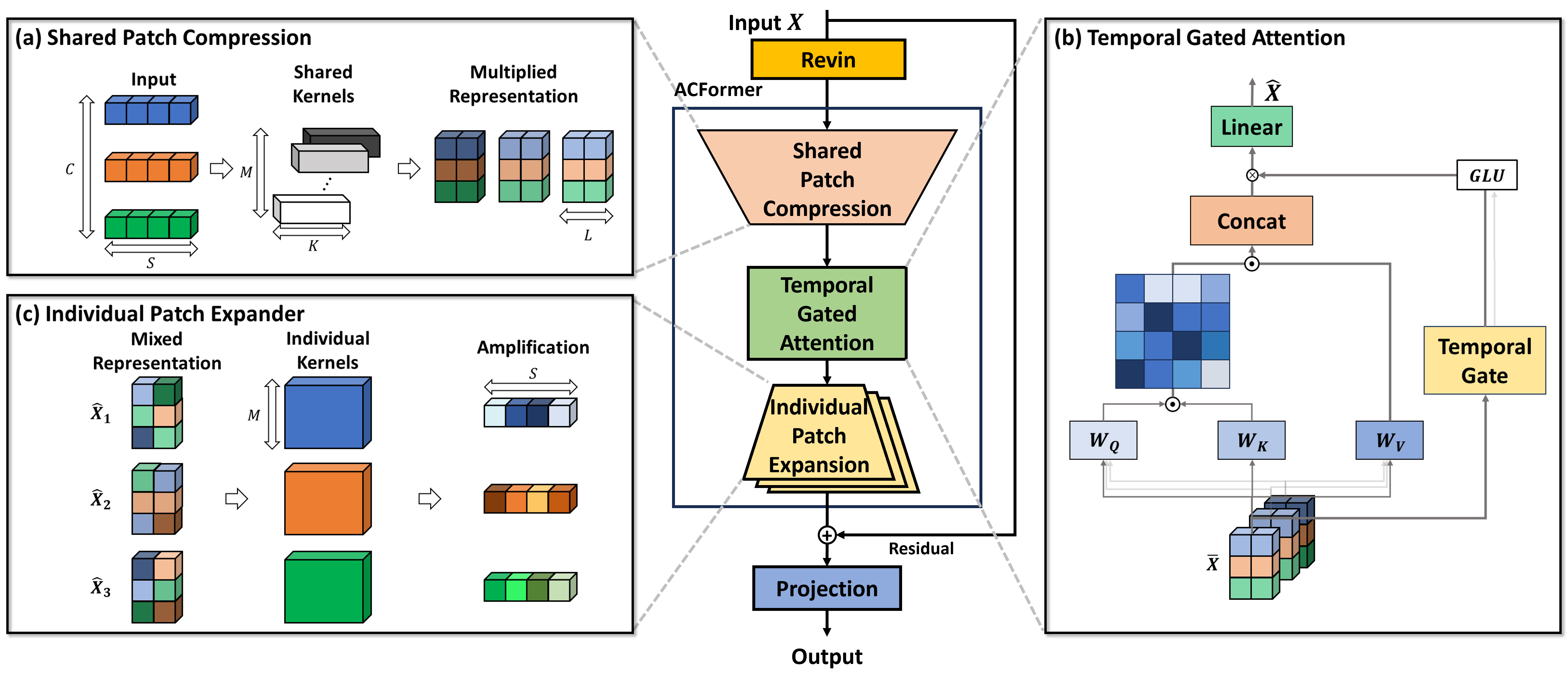}
    \caption{The overall framework of ACFormer, illustrating the flow from Shared Patch Compression through Temporal Gated Attention to Independent Patch Expansion.}
    \label{fig:Framework}
\end{figure*}

\section{Methodology}

\subsection{Shared Patch Compression}

We used the reversible instance normalization methods as pre-processing techniques. Each instance was processed by subtracting its mean and dividing by the standard deviation \cite{RevIN}. This ensures that the subsequent convolutional kernels operate on a stable, zero-centered signal, allowing the model to focus on capturing structural patterns.

To integrate the inductive bias of CNNs into a channel-wise attention framework, we introduce Shared Patch Compression. This module performs simultaneous temporal down-sampling and multi-representation encoding to capture robust local features while reducing complexity. Treating each channel as an independent sequence $X^p\in\mathbb{R}^{S\times 1\times C}$, the operation is formulated as follows:
\begin{equation}
    \bar{X}_m=Conv(X^p;W_{m}^{CP},T)
\end{equation}
where $W_m^{CP}\in \mathbb{R}^K$ is the $m$-th convolutional kernel ($m\in{1,\dots,M}$) and $T$ is the stride size. Output $\bar{X}\in\mathbb{R}^{L\times M\times C}$ provides $M$ with distinct temporal representations per channel, serving as the head for multi-head attention. The compressed sequence length $L=\lfloor (S-K)/T+1\rfloor$ is maintained without zero-padding to avoid artifacts.

This strategy enhances channel-wise attention in two ways: (1) Unlike linear embeddings that treat sequences as flat vectors, the convolutional stride aggregates temporal dependencies. This filters high-frequency information and distills stable global patterns, such as trends and seasonality, preserving structural integrity of time series data for inter-channel modeling. (2) Convolutional kernels act as learnable extractors that map non-stationary signals into diverse latent views. By simultaneously capturing rapid fluctuations and gradual shifts, this approach allows the multi-headed  attention mechanism to reconcile correlations across varying temporal scales.

\subsection{Temporal Gated Attention} 

To model inter-variable correlations, we adapted the channel-wise attention framework with modifications optimized for auto-convolution. Standard attention mechanisms often project concatenated head representations, thereby incurring in significant computational overhead. On the other hand, Shared Patch Compression distills distinct features into separate heads; therefore, we applied linear projections to each head independently. The each head of query $(Q_m)$, key $(K_m)$, and value $(V_m)$ head is respectively computed as:

\begin{equation}
\begin{split}
    Q_m = W_Q\cdot \bar{X}_m+b_Q \\
    K_m = W_K\cdot \bar{X}_m+b_K  \\
    V_m = W_V\cdot \bar{X}_m+b_V    
\end{split}
\end{equation}
where $Q_m,K_m,V_m\in\mathbb{R}^{L\times C}$ and $W_Q, W_K, W_V\in\mathbb{R}^{L\times L}$ are weight matrices, and $b_Q,b_K, b_V\in\mathbb{R}^{L}$ are the biases. By applying head-independent projection, the parameter requirements are reduced by a factor of $M^2$ compared to standard multi-head attention, thereby enhancing efficiency without sacrificing diversity. The attention scores are computed as $A=Q^TK/\sqrt{f}\in \mathbb{R}^{C\times C}$, and then multiplied by $V$ to produce the self attention output $H_A\in\mathbb{R}^{L\times M\times C}$. 

Although channel-wise attention captures inter-variable correlations, matrix multiplication along the time axis can disregard local temporal dependencies. To mitigate this, we introduce a Temporal Gate that ensures that each time step is informed by its neighbors, thereby further expanding the receptive field.  By reshaping the input into $\bar{X}^p\in\mathbb{R}^{L\times 1\times (C\cdot H)}$, the gate values and final gated results are obtained as follows: 
\begin{equation}
\begin{split}
    G = Conv(\bar{X};W_{G})\\
    H_G = H_A\cdot GLU(G)
\end{split}
\end{equation}
where $W_{G}\in\mathbb{R}^{K'\times 2}$ represents a convolution kernel with a kernel size of $K'$ and $H_G\in\mathbb{R}^{L\times M\times C}$ is the gated output. The convolution layer utilizes two kernels to output two different representations for $GLU$ activation. We employed isomorphic convolutions with the padding size $\lfloor(K'-1)/2\rfloor$ to preserve the input dimensions. Finally, $H_G$ concatenates along the head dimension and is processed using feedforward networks and batch normalization.  

\begin{table*}[ht]
\centering
\small
\setlength{\tabcolsep}{2.2pt} 
\renewcommand{\arraystretch}{1.1}

\caption{Multivariate long-term forecasting results (MSE and MAE) across multiple benchmarks ($S=96, P \in \{96, 192, 336, 720\}$).}
\label{tab:long_term_average}
\resizebox{\textwidth}{!}{
    \begin{tabular}{c|cc|cc|cc|cc|cc|cc|cc|cc|cc}
    
    \toprule[1pt]
    \multirow{2}{*}{Models} & 
    \multicolumn{2}{c|}{ACFormer} & 
    \multicolumn{2}{c|}{TimeMixer++} & 
    \multicolumn{2}{c|}{TimePro} & 
    \multicolumn{2}{c|}{Amplifier} & 
    \multicolumn{2}{c|}{iTransformer} & 
    \multicolumn{2}{c|}{ModernTCN} & 
    \multicolumn{2}{c|}{PatchTST} & 
    \multicolumn{2}{c|}{TimesNet} & 
    \multicolumn{2}{c}{DLinear} \\
    & 
    \multicolumn{2}{c|}{(Ours)} & \multicolumn{2}{c|}{(2025)} & \multicolumn{2}{c|}{(2025)} & \multicolumn{2}{c|}{(2025)} & \multicolumn{2}{c|}{(2024)} & \multicolumn{2}{c|}{(2024)} & \multicolumn{2}{c|}{(2023)} & \multicolumn{2}{c|}{(2023)} & \multicolumn{2}{c}{(2023)}  \\
    
    \midrule
    Metric & MSE & MAE & MSE & MAE & MSE & MAE & MSE & MAE & MSE & MAE & MSE & MAE & MSE & MAE & MSE & MAE & MSE & MAE\\
    
    \midrule
    {ecl}&
    \textbf{0.167}&\textbf{0.257}&0.190&0.274&0.180&0.264&0.178&0.264&0.178&\underline{0.263}&\underline{0.174}&0.270&0.194&0.270&0.201&0.293&0.212&0.288\\
    \midrule
    {etth1}&
    \textbf{0.447}&\textbf{0.424}&0.460&0.436&\underline{0.449}&0.438&0.463&0.434&0.453&0.440&0.483&0.451&0.454&\underline{0.431}&0.476&0.455&0.458&0.436\\
    \midrule
    {etth2}&
    \textbf{0.365}&\textbf{0.390}&0.379&0.398&0.371&\underline{0.391}&0.372&0.393&0.386&0.403&\underline{0.368}&0.394&0.380&0.396&0.392&0.406&0.432&0.435\\
    \midrule
    {ettm1}&
    \textbf{0.378}&\textbf{0.377}&\underline{0.381}&0.384&0.392&0.387&0.390&0.382&0.389&0.386&0.400&0.410&\underline{0.381}&\underline{0.379}&0.464&0.432&0.396&0.388\\
    \midrule
    {ettm2}&
    \textbf{0.273}&\textbf{0.314}&0.289&0.325&0.282&0.322&\underline{0.274}&\underline{0.316}&0.281&0.320&0.290&0.327&0.275&0.318&0.300&0.330&0.287&0.331\\
    \midrule
    {sol}&
    \textbf{0.228}&\textbf{0.229}&0.235&0.238&0.335&0.305&\underline{0.230}&\underline{0.234}&0.237&\underline{0.234}&0.244&0.254&0.255&0.251&0.278&0.259&0.335&0.314\\
    \midrule
    {traffic}&
    0.468&\textbf{0.274}&0.495&\underline{0.278}&\underline{0.450}&\textbf{0.274}&0.470&0.294&\textbf{0.441}&0.281&0.476&0.294&0.698&0.316&0.654&0.326&0.646&0.355\\
    \midrule
    {weather}&
    \textbf{0.238}&\textbf{0.261}&0.248&0.267&0.278&0.292&0.242&0.264&0.290&0.302&\underline{0.241}&0.267&0.243&\underline{0.263}&0.264&0.283&0.272&0.291\\
    
    \bottomrule[1pt]
    \end{tabular}
}
\end{table*}

\subsection{Individual Patch Expansion}

The final stage of the ACFormer architecture is the individual patch expansion layer, which serves as a decoder within the Auto-Convolution framework. Althoguh the initial compression layer uses shared kernels to extract universal temporal patterns, the reconstruction of high-frequency components requires a more granular approach. As high-frequency fluctuations and seasonal patterns are often unique to each variable, a shared kernel fails to capture this inter-channel heterogeneity.

To address this, we employ Channel-Independent Transposed Convolutions to map hidden states back to the original sequence length. Unlike the compression layer, the weights of this expansion layer are not shared across channels. The operation for each channel $c$ is defined as:

\begin{equation} 
X^{amp}_c = \text{TransposedConv}(H_c; W_c^{EP}, T) 
\end{equation}

where $H_c\in\mathbb{R}^{L\times M}$ are single channel values with multiplied representation from attention layer outputs, $W_c^{EP}\in \mathbb{R}^{K\times M}$ denotes the channel-specific expansion kernel, and $T$ is the stride size matching the encoder. This kernel utilizes $M$ hidden representations to reconstruct the fine-grained temporal details. By allowing each channel to maintain its own unique expansion parameters, the model performs high-frequency amplification, recovering the non-linear nuances typically lost in purely linear projection-based models.

Following reconstruction of the fine-grained details, the amplified features $X^{amp}$ are integrated with the base signal. To extrapolate with the amplified information, a linear projection layer is applied. Finally, the RevIN denormalization step is applied to restore the instance-specific mean and variance removed during the initial compression stage.

\begin{equation} 
\hat{Y} = \text{Denorm}(\text{Linear}(X + X^{amp})) 
\end{equation}

The model is trained to minimize the MAE between the prediction and ground truth:

\begin{equation}
    \mathcal{L} = \sum_c|Y_c-\hat{Y_c|}
\end{equation}

\section{Experiments}
\subsection{Experiment Setup}

\textbf{Baselines} To evaluate the proposed ACFormer performance, we selected SOTA architectures in Multivariate TSF domain across four distinct categories: (1) \textbf{Transformer-based models}: iTransformer \cite{iTransformer}, and PatchTST \cite{PatchTST}; (2) \textbf{Linear-based models}: TimeMixer++ \cite{TimeMixerPP}, Amplifier \cite{Amplifier}, and DLinear \cite{DLinear}; (3) \textbf{CNN-based models}: ModernTCN \cite{ModernTCN}, and TimesNet \cite{TimesNet}; and (4) \textbf{State Space-based}: TimePro \cite{TimePro}. The models were evaluated on the validation set after each epoch using MAE. The weights corresponding to the lowest validation loss were then evaluated on the test set using the mean squared error (MSE) and MAE.

\textbf{Datasets} We evaluated our model on six large-scale, real-world benchmark datasets: (1) \textbf{Electricity Consuming Load (ECL)} \cite{Autoformer} data comprise hourly electricity consumption with 321 variables; (2) \textbf{Electricity transformer temperature (ETT)} \cite{Autoformer}, which includes seven variables resampled using two criteria: hourly sampled data (ETTh1 and ETTh2) and 15-m sampled data (ETTm1 and ETTm2); (3) \textbf{Solar energy} \cite{LTSNet} cite which includes 137 variables sampled every 10 mins; (4) \textbf{Traffic} \cite{Autoformer} data, which describe the occupancy rates of lanes using 862 variables; (5) \textbf{Weather} \cite{Autoformer}, which includes 21 variables collected every 10 min from a weather station; and (6) \textbf{PEMS} \cite{SCINet} data, which contains public traffic network data sampled every 5-min. All data were split chronologically. The ETT and PEMS datasets utilized a 6:2:2 ratio for training, validation, and testing, whereas others used a 7:1:2 ratio. 

\begin{table*}[th]
\centering
\small
\setlength{\tabcolsep}{2.2pt} 
\renewcommand{\arraystretch}{1.1}

\caption{Multivariate short-term forecasting results (MSE and MAE) for various prediction horizons.}
\resizebox{\textwidth}{!}{
    \begin{tabular}{c|cc|cc|cc|cc|cc|cc|cc|cc|cc}
    
    \toprule[1pt]
    \multirow{2}{*}{Models} & 
    \multicolumn{2}{c|}{ACFormer} & 
    \multicolumn{2}{c|}{TimeMixer++} & 
    \multicolumn{2}{c|}{TimePro} & 
    \multicolumn{2}{c|}{Amplifier} & 
    \multicolumn{2}{c|}{iTransformer} & 
    \multicolumn{2}{c|}{ModernTCN} & 
    \multicolumn{2}{c|}{PatchTST} & 
    \multicolumn{2}{c|}{TimesNet} & 
    \multicolumn{2}{c}{DLinear} \\
    & 
    \multicolumn{2}{c|}{(Ours)} & \multicolumn{2}{c|}{(2025)} & \multicolumn{2}{c|}{(2025)} & \multicolumn{2}{c|}{(2025)} & \multicolumn{2}{c|}{(2024)} & \multicolumn{2}{c|}{(2024)} & \multicolumn{2}{c|}{(2023)} & \multicolumn{2}{c|}{(2023)} & \multicolumn{2}{c}{(2023)}  \\
    
    \midrule
    Metric & MSE & MAE & MSE & MAE & MSE & MAE & MSE & MAE & MSE & MAE & MSE & MAE & MSE & MAE & MSE & MAE & MSE & MAE\\
    
    \midrule
    {pems03}&
    \textbf{0.117}&\textbf{0.222}&0.240&0.328&0.331&0.375&\underline{0.131}&\underline{0.236}&0.151&0.253&0.206&0.303&0.208&0.294&0.154&0.247&0.262&0.343\\
    \midrule
    {pems04}&
    \textbf{0.114}&\textbf{0.222}&0.268&0.354&0.398&0.416&0.139&0.246&\underline{0.122}&\underline{0.230}&0.205&0.301&0.282&0.342&0.131&0.237&0.276&0.349\\
    \midrule
    {pems07}&
    \textbf{0.106}&\textbf{0.197}&0.263&0.356&0.172&0.270&\underline{0.114}&0.214&0.114&\underline{0.208}&0.206&0.310&0.221&0.294&0.126&0.219&0.319&0.366\\
    \midrule
    {pems08}&
    \textbf{0.141}&\textbf{0.229}&0.269&0.341&0.227&0.304&\underline{0.177}&0.264&0.198&0.272&0.274&0.345&0.249&0.311&0.185&\underline{0.260}&0.340&0.364\\
    \midrule
    {etth1}&
    \textbf{0.304}&\textbf{0.343}&0.310&0.350&0.336&0.374&\underline{0.307}&0.346&0.317&0.356&0.308&\underline{0.345}&0.308&0.346&0.342&0.369&0.311&0.346\\
    \midrule
    {etth2}&
    \underline{0.181}&\underline{0.262}&0.184&0.263&0.188&0.265&0.182&0.263&0.190&0.269&\underline{0.181}&0.265&\textbf{0.177}&\textbf{0.259}&0.200&0.279&0.183&0.263\\
    
    \bottomrule[1pt]
    \end{tabular}
}
\label{tab:short_term_average}
\end{table*}

\begin{figure}[ht]
  \centering
  \includegraphics[width=0.95\linewidth]{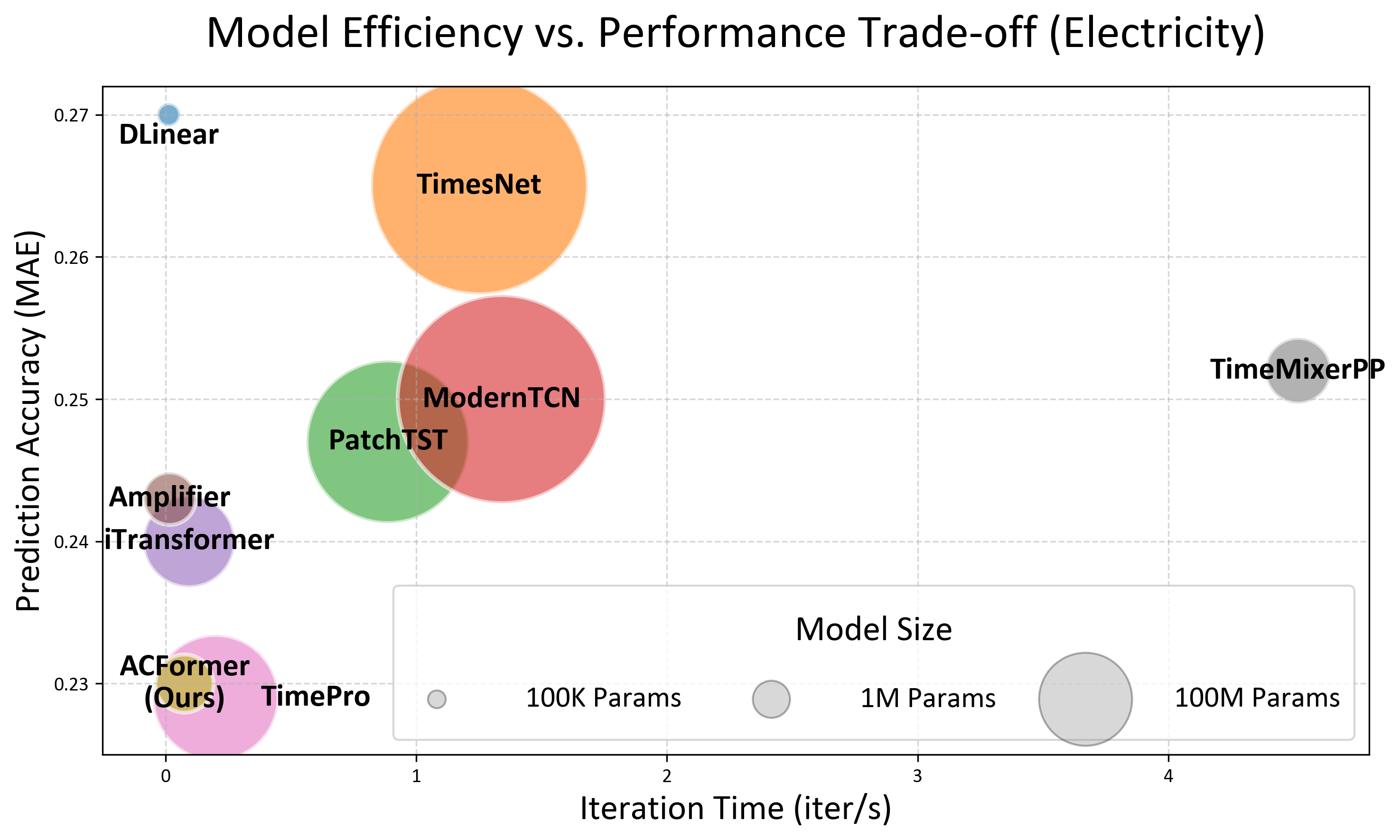}
  \caption{Model efficiency(x-axis) and performance (y-axis) trade-off on the ECL dataset ($S=96, P=96$).}
  \label{fig:model_efficiency}
\end{figure}

\textbf{Implementation details} All experiments were conducted in a Google Colab cloud-computing environment using an NVIDIA L4 GPU for training. Training was performed for 10 epochs using the Adam optimizer with a batch size of 32 and an early stopping patience of three epochs. A uniform initial learning rate of 0.001 was applied to all models and adjusted using a scheduler after each epoch. The input window was fixed at $S=96$ for both long- and short-term forecasting, whereas the prediction length varied by $P=\{96, 192, 336, 720\}$ for long-term, and $\{12, 24, 48, 96\}$ or $\{12, 24, 48, 60\}$ for short-term forecasting.  

\subsection{Experimental Results}
The multivariate long- and short-term forecasting results are summarized in Tables \ref{tab:long_term_average} and \ref{tab:short_term_average}, respectively. Across both tasks, ACFormer consistently outperforms all baselines, achieving new SOTA results. 

\textbf{Performance and Robustness} Significant performance gains are observed in comparison with to iTransformer, particularly on non-stationary datasets. For instance, on the Weather dataset, which is characterized by high non-stationarity, ACFormer demonstrates a substantial improvement in terms of predictive accuracy. This validates our core proposition: while channel-wise attention excels in relational modeling, the integration of convolutional layers allows ACFormer to capture the intricate non-linear patterns and high-frequency fluctuations that linear-based models often neglect. 

\textbf{Computational Efficiency} Beyond accuracy, ACFormer exhibits superior efficiency. As illustrated in Figure \ref{fig:model_efficiency}, ACFormer achieves superior results with a significantly smaller parameter footprint than other Transformer-based models, matching the efficiency of MLP-based architectures. This balance between high-fidelity feature extraction and low computational overhead makes ACFormer a highly practical solution for real-world dataset.

\begin{table}[ht]
\centering
\small
\caption{Ablation study results on the ETTh1, Solar, and Weather datasets}
\begin{tabular}{cc|cc|cc|cc}

\toprule[1pt]
\multicolumn{2}{c|}{Model} & 
\multicolumn{2}{c|}{W/O Attention} & 
\multicolumn{2}{c|}{W/O Gate} & 
\multicolumn{2}{c}{ACFormer}
\\

\midrule
\multicolumn{2}{c|}{Metric} & MSE & MAE & MSE & MAE & MSE & MAE \\

\midrule
\multirow{4}{*}{\rotatebox[origin=c]{90}{etth1}}
&96&0.385&\textbf{0.386}&0.383&\textbf{0.386}&\textbf{0.382}&0.387\\
&192&0.441&0.417&\textbf{0.438}&\textbf{0.416}&\textbf{0.438}&\textbf{0.416}\\
&336&0.485&0.437&0.488&0.439&\textbf{0.479}&\textbf{0.436}\\
&720&0.489&0.461&0.497&0.466&\textbf{0.488}&\textbf{0.459}\\
\midrule
\multirow{4}{*}{\rotatebox[origin=c]{90}{sol}}
&96&0.266&0.264&0.191&0.208&\textbf{0.189}&\textbf{0.205}\\
&192&0.306&0.283&0.249&0.240&\textbf{0.221}&\textbf{0.224}\\
&336&0.345&0.302&0.264&0.250&\textbf{0.244}&\textbf{0.241}\\
&720&0.345&0.296&0.260&0.250&\textbf{0.256}&\textbf{0.248}\\
\midrule
\multirow{4}{*}{\rotatebox[origin=c]{90}{weather}}
&96&0.165&0.202&\textbf{0.150}&0.188&\underline{0.150}&\textbf{0.187}\\
&192&0.209&0.243&0.209&0.243&\textbf{0.201}&\textbf{0.239}\\
&336&0.265&0.284&0.262&\textbf{0.283}&\textbf{0.261}&0.284\\
&720&0.345&0.336&\textbf{0.339}&0.334&\textbf{0.339}&\textbf{0.333}\\

\bottomrule[1pt]
\end{tabular}
\label{tab:ablation_study}
\end{table}

\subsection{Model Analysis}

\subsubsection{Ablation study}
To evaluate the contribution of each component within ACFormer, we conducted an ablation study across the ETTh1, Solar, and Weather datasets, using the same experimental settings as in the long-term forecasting task. We compared the full ACFormer architecture with two variants: (1)\textbf{W/O Gate} to evaluate the impact of the temporal gate module by removing it from the attention layer; and (2) \textbf{W/O Attention} to assesses the performance of a purely convolutional auto-encoder consisting only of the Shared Patch Compression and Individual Patch Expansion modules. The results are summarized in Table \ref{tab:ablation_study}

Empirical evidence suggests that while the model performs adequately without a temporal gate, performance degrades when the gate is absent for datasets with strong periodicity such as Solar. This confirms that the temporal gate is essential for preserving local temporal dependencies that are often lost during matrix multiplication.

A striking result is observed for the W/O Attention variant. On the Weather dataset, the purely convolutional framework performed sufficiently well, outperforming existing baselines, while incurring in a lower computational burden than DLinear. This finding underscores the superior ability of convolutional layers to extract and model non-linear information from raw, non-stationary time series.

\subsubsection{Comparison with iTransformer}

\begin{figure}[h]
    \centering
    \includegraphics[width=0.95\linewidth]{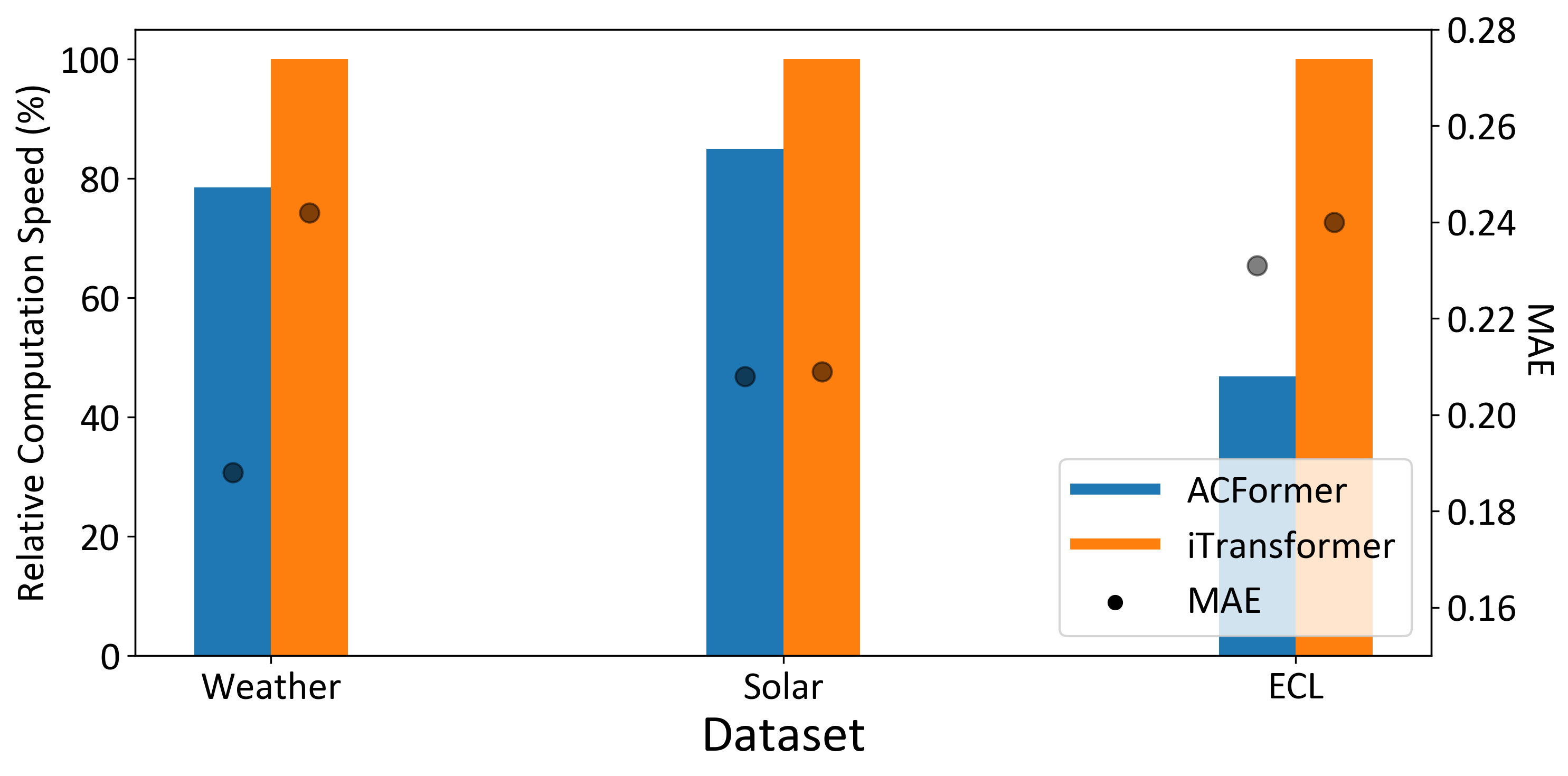}
    \caption{Relative iteration time of ACFormer (Blue) and iTransformer(Orange) across the Weather, Solar, and Electricity datasets ($S=96, P=96$).}
    \label{fig:relative_time}
\end{figure}

To investigate the difference of ACFormer and iTransformer, systemical comparison is conducted. ACFormer consistently outperformed iTransformer across diverse benchmarks while demonstrating superior computational efficiency. As shown in Figure \ref{fig:relative_time}, we measured the relative iteration time ($S=96, P=96$) for the Weather, Solar, and Electricity datasets. For a more detailed explanation, a value of 80 for ACFormer for the Weather dataset indicates that ACFormer required 80\% of the iteration time required by iTransformer. These results suggests that ACFormer significantly reduces the computational burden, most notably on the Electricity dataset, for which it achieves a 50\% reduction in iteration time compared with iTransformer while simultaneously improving the prediction accuracy.

\begin{figure}[h]
    \centering
    \subfloat[\centering Weather - iTransformer]{{\includegraphics[width=.5\linewidth]{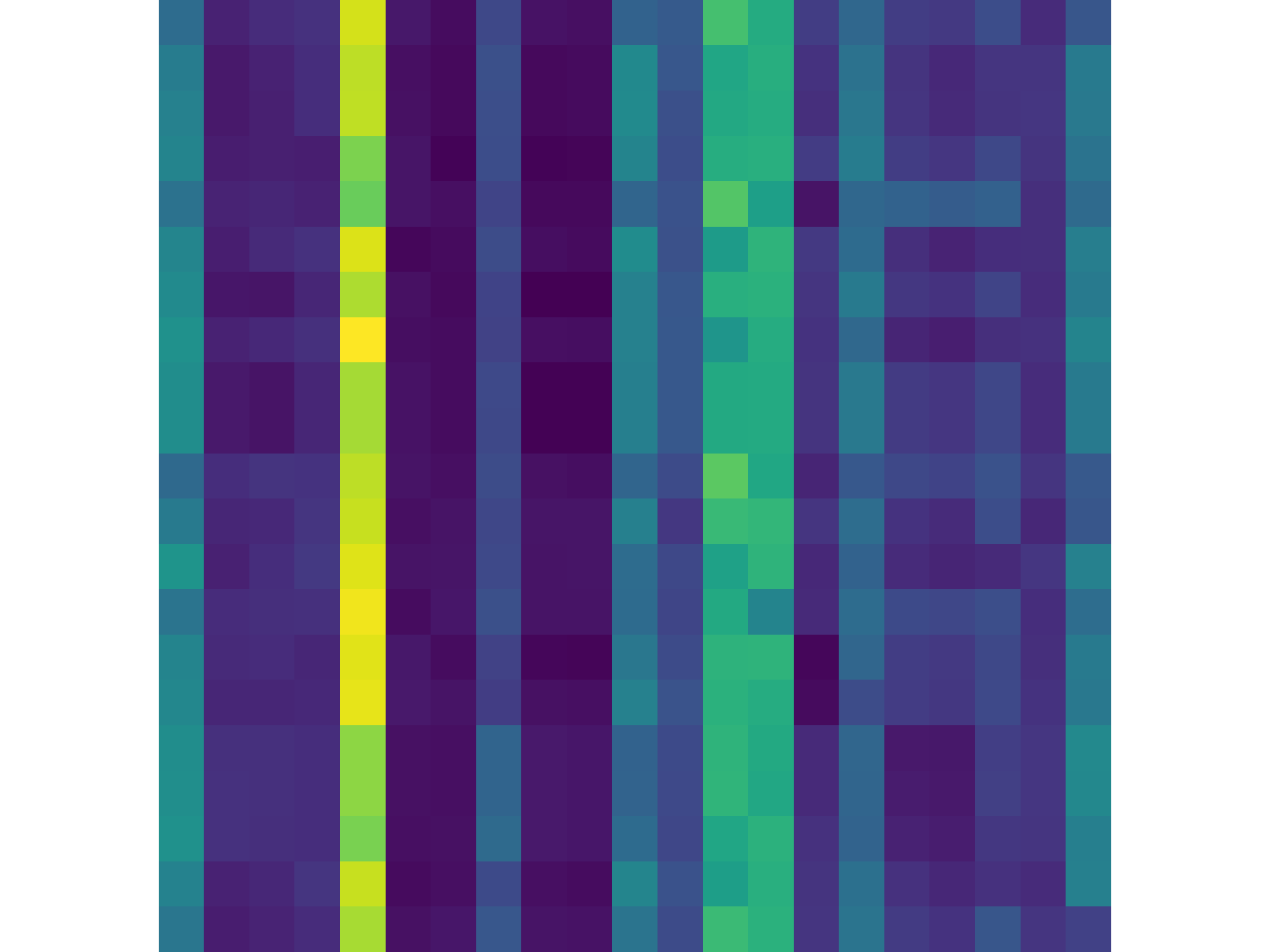} }}%
    \subfloat[\centering Solar - iTransformer]{{\includegraphics[width=.5\linewidth]{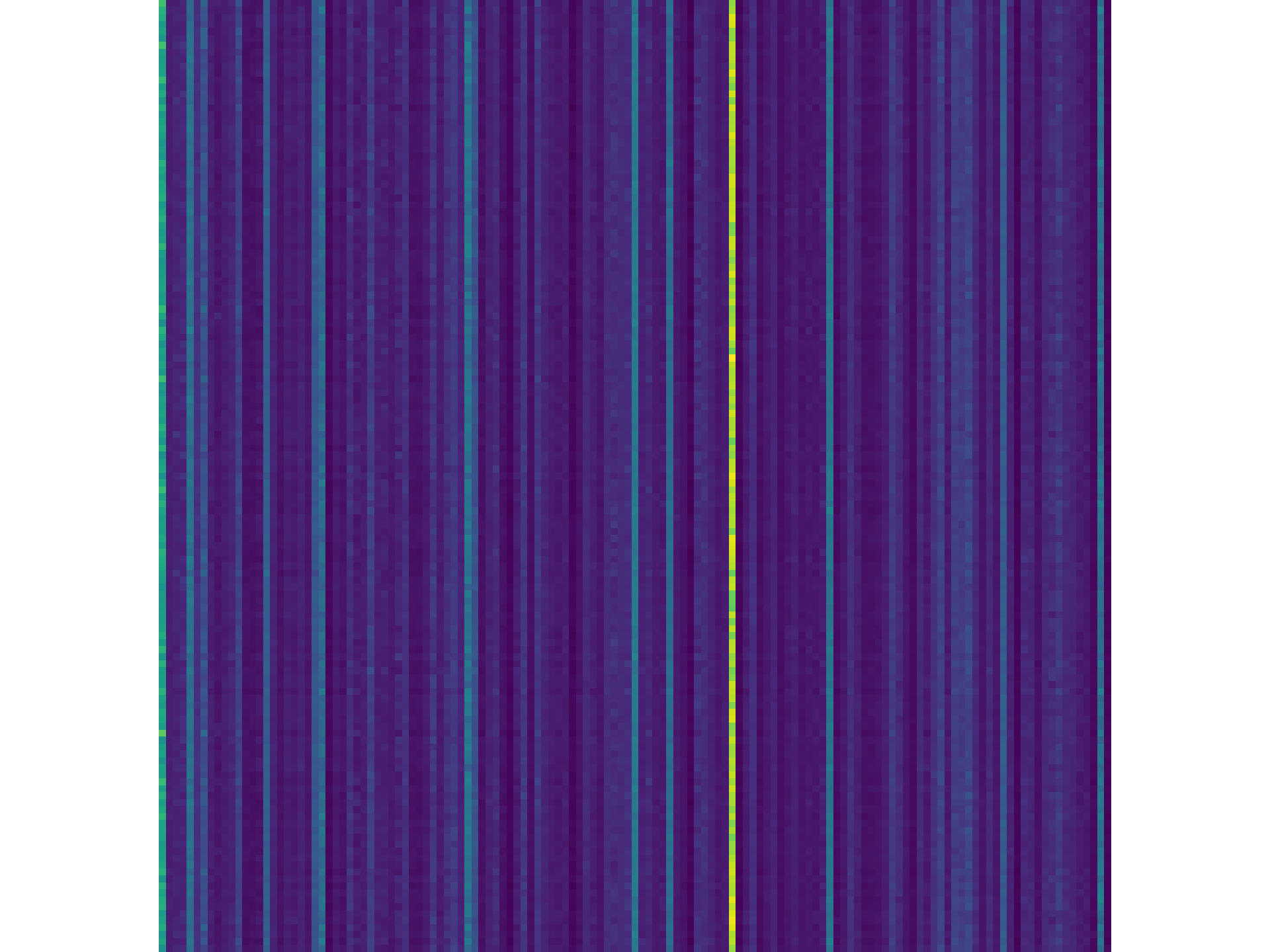} }}%

    \vskip\baselineskip

    \subfloat[\centering Weather - ACFormer]{{\includegraphics[width=.5\linewidth]{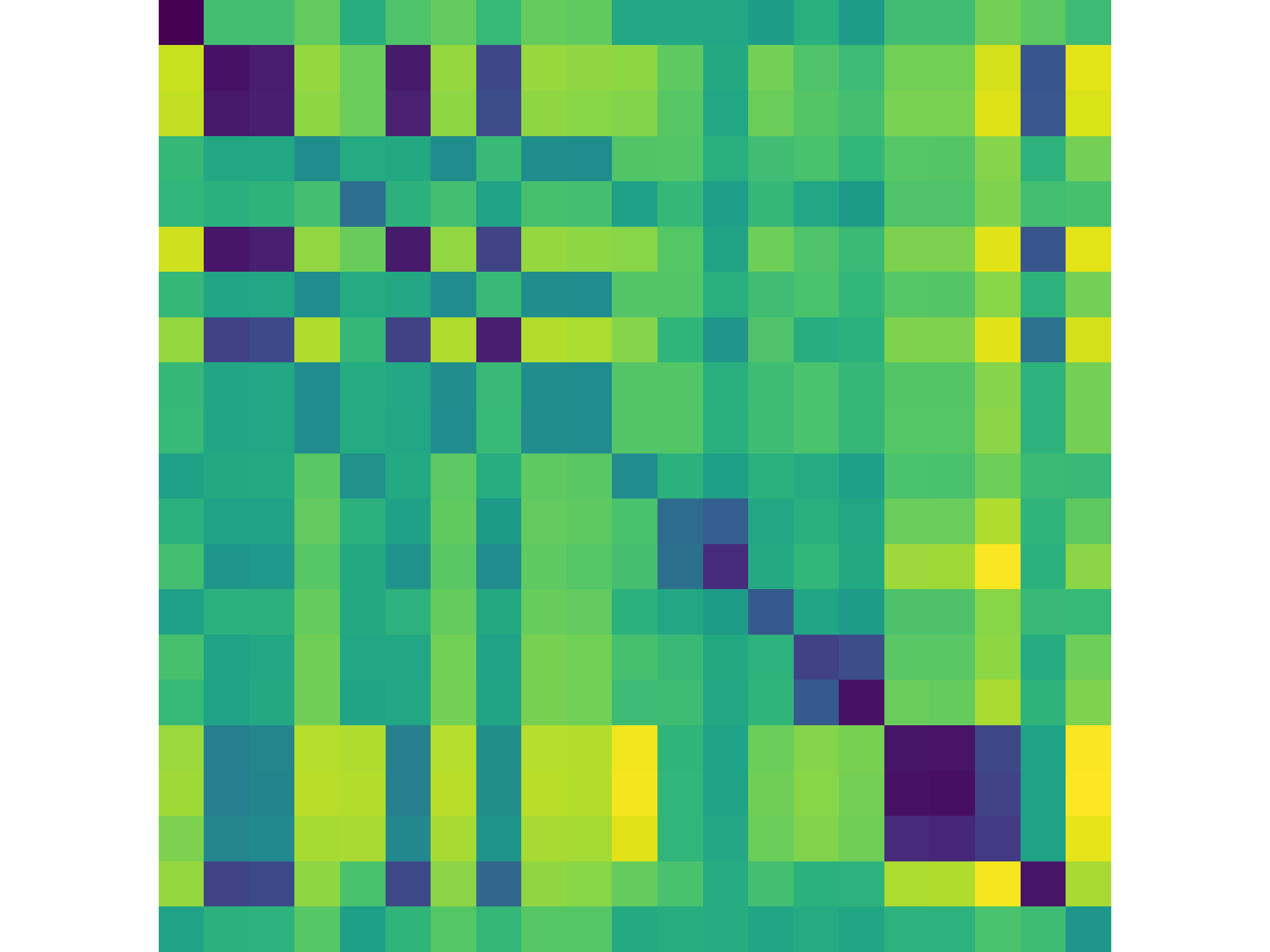} }}%
    \subfloat[\centering Solar - ACFormer]{{\includegraphics[width=.5\linewidth]{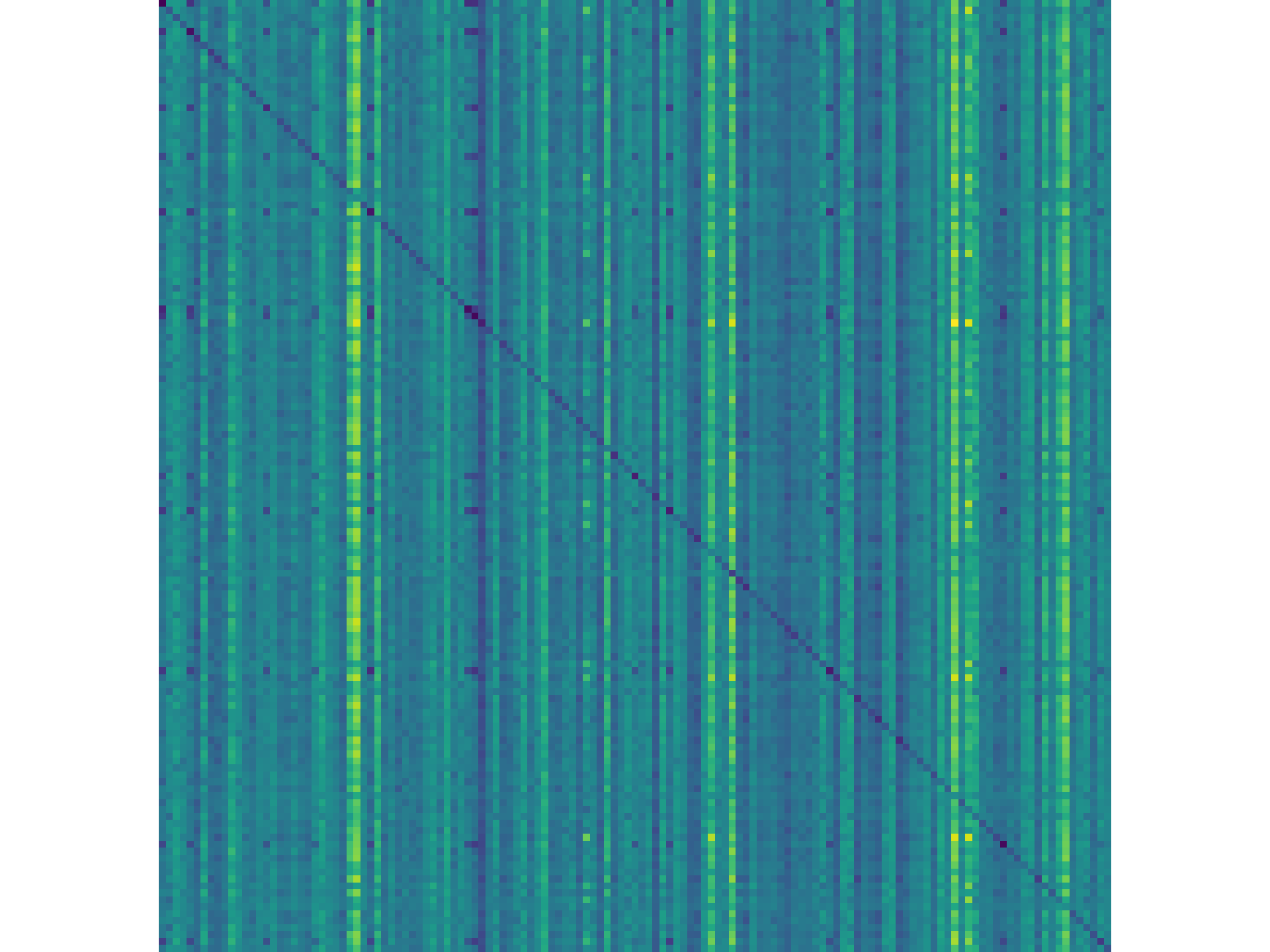} }}%
    
    \vskip\baselineskip

    \subfloat[\centering Weather - Correlation]{{\includegraphics[width=.5\linewidth]{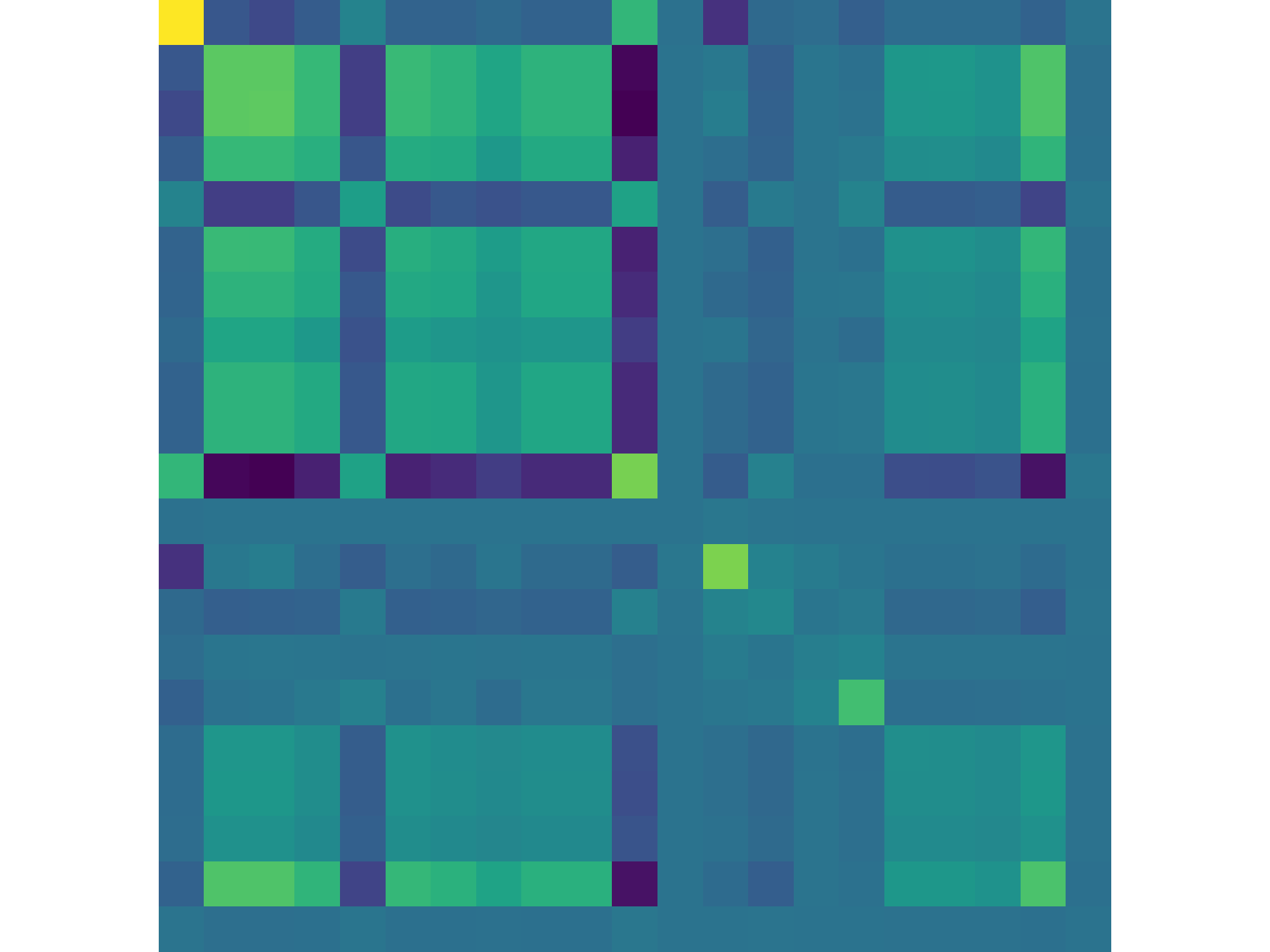} }}%
    \subfloat[\centering Solar - Correlation]{{\includegraphics[width=.5\linewidth]{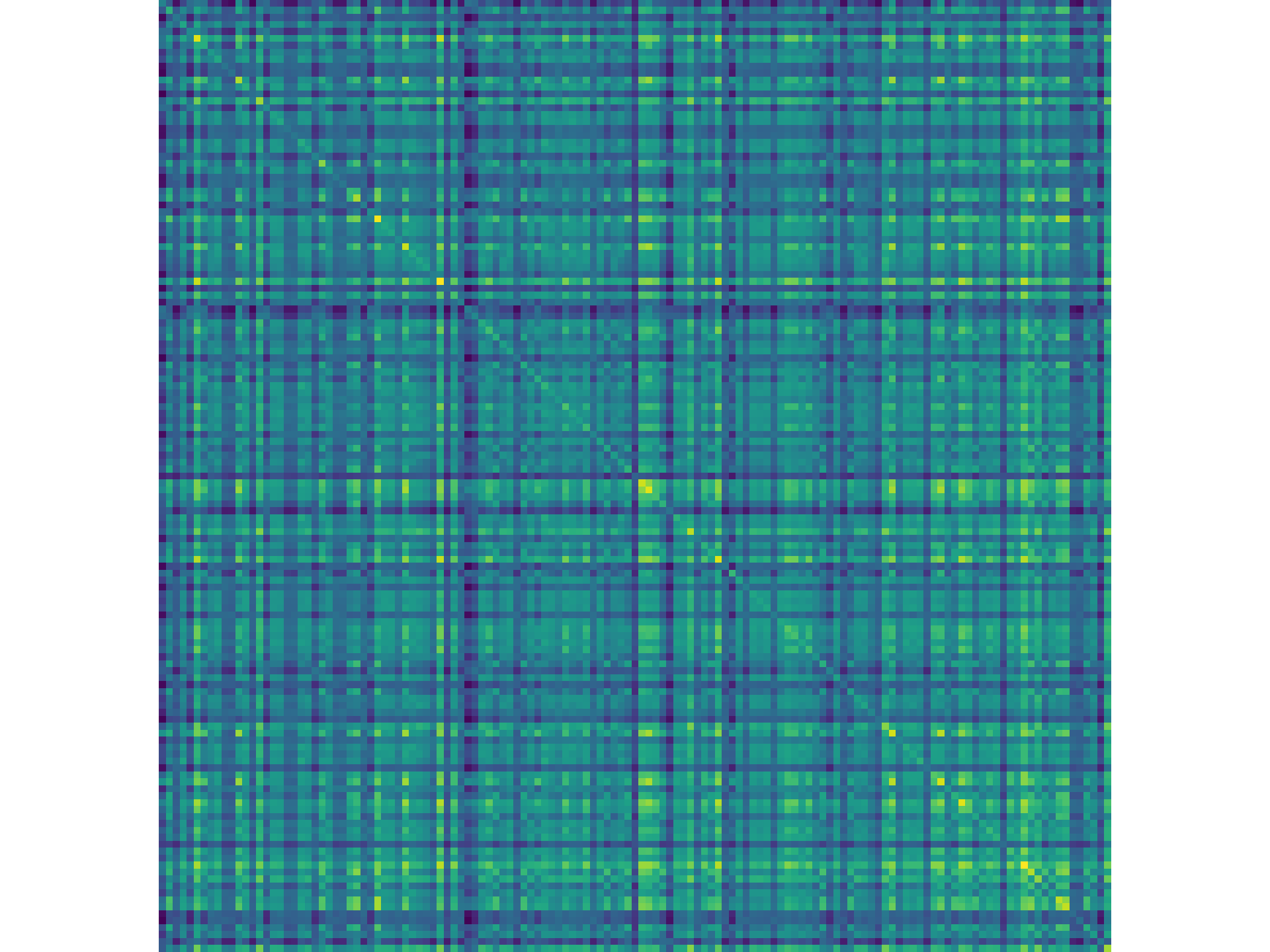} }}%

    \caption{Learned channel-wise attention maps and ground-truth channel correlations}%
    \label{fig:Attention_comparison}%
\end{figure}

To investigate the underlying reasons for this performance boost, we visualized the channel-wise attention values from the final layers of both models, as shown in Figure \ref{fig:Attention_comparison}. While previous studies suggests that channel-wise attention should mirror the correlation between variables, we observed that iTransformer's attention maps fail to maintain this resemblance on non-stationary datasets such as the Weather data. By contrast, ACFormer’s attention layers exhibit a much higher resemblance to the actual data correlations. By incorporating fine-grained temporal features through convolutional dynamics, ACFormer learns more meaningful inter-variable relationships, which directly contributes to its superior forecasting accuracy.

\section{Conclusion}
In this study, we have conducted in depth analysis of convolutional layers in TSF domain to resolve the limitation of linear models. Through a novel individual receptive field analysis, we demonstrate that convolutional layers effectively identify "pivot channels," mirroring explicit channel-wise attention while providing superior robustness to non-linear fluctuations.

Building on these insights, we propose ACFormer, an architecture that reconciles the efficiency of linear projections with the non-linear feature-extraction power of convolutions. By utilizing Shared Patch Compression to distill universal patterns and an Independent Patch Expansion layer to amplify variable-specific high-frequency nuances, combined with channel-wise attention, ACFormer preserves the structural integrity of complex sequences. Extensive experiments across real-world benchmarks demonstrate that ACFormer achieves state-of-the-art performance while maintaining higher computational efficiency than iTransformer.

\newpage
\bibliography{ACFormer}
\bibliographystyle{icml2026}

\newpage
\appendix
\onecolumn
\section{Experimental details}

\subsection{Datasets}
We evaluated the ACFormer's performance on ten real-world datasets for a comparative analysis. Details of each dataset are provided next: (1) ECL (Electricity Consuming Load): This dataset \cite{Autoformer} contains hourly electricity consumption data for 321 clients. (2) ETT (Electricity Transformer Temperature): From July 2016 to July 2018, this dataset \cite{Autoformer} includes seven factors related to electricity transformers. It is divided into four subsets: ETTh1 and ETTh2 (hourly data), and ETTm1 and ETTm2 (15-min data). (3) Solar-Energy: This dataset \cite{LTSNet} contains solar power production from 137 photovoltaic (PV) plants in 2006, with measurements taken every 10 min. (4) Traffic: The traffic dataset \cite{Autoformer} contains hourly road occupancy rates from 862 sensors on San Francisco Bay area freeways, collected between January 2015 and December 2016. (5) Weather: The weather dataset \cite{Autoformer} includes 21 meteorological factors collected every 10 min from the Weather Station of the Max Planck Biogeochemistry Institute throughout 2020. (6) PEMS: This dataset \cite{LTSNet} contains public traffic network data for California. It is sampled in 5-min windows and use four subsets named as PEMS03, PEMS04, PEMS07, and PEMS08. Table \ref{tab:Data_summ} summarizes these datasets.

\begin{table*}[th]
\centering
\caption{Summary of dataset statistics, including variable dimensions, sample sizes, and sampling frequencies.}
\resizebox{0.7\textwidth}{!}{
  \begin{tabular}{c|ccc}
  \bottomrule[1pt]
  Model & Dimension & \makecell{Dataset size \\  (Train, Val, Test)} & Frequency \\
  \hline
  ECL & 321 & (18317, 2633, 5261) & Hourly \\
  ETTm & 7 & (34465, 11521, 11521) & 15 mins \\
  ETTh & 7 & (8545, 2881, 2881) & Hourly \\
  Solar Energy & 127 & (36601, 5161, 10417) & 10 mins \\
  Traffic & 862 & (12185, 1757, 3509) & Hourly \\
  Weather & 21 & (36792, 5271, 10540) & 10 mins \\
  PEMS03 & 358 & (15617, 5135, 5135) & 5 mins \\
  PEMS04 & 307 & (10172, 3375, 3375) & 5 mins \\
  PEMS07 & 883 & (16911, 5622, 5622) & 5 mins \\
  PEMS08 & 170 & (10690, 3548, 3548) & 5 mins \\
  \toprule[1pt]

  \end{tabular}
}
\label{tab:Data_summ}
\end{table*}


\subsection{Individual Receptive Fields}
\label{appendix:individual_receptive_field}

\begin{figure}[H]
    \centering
    \includegraphics[width=0.85\linewidth]{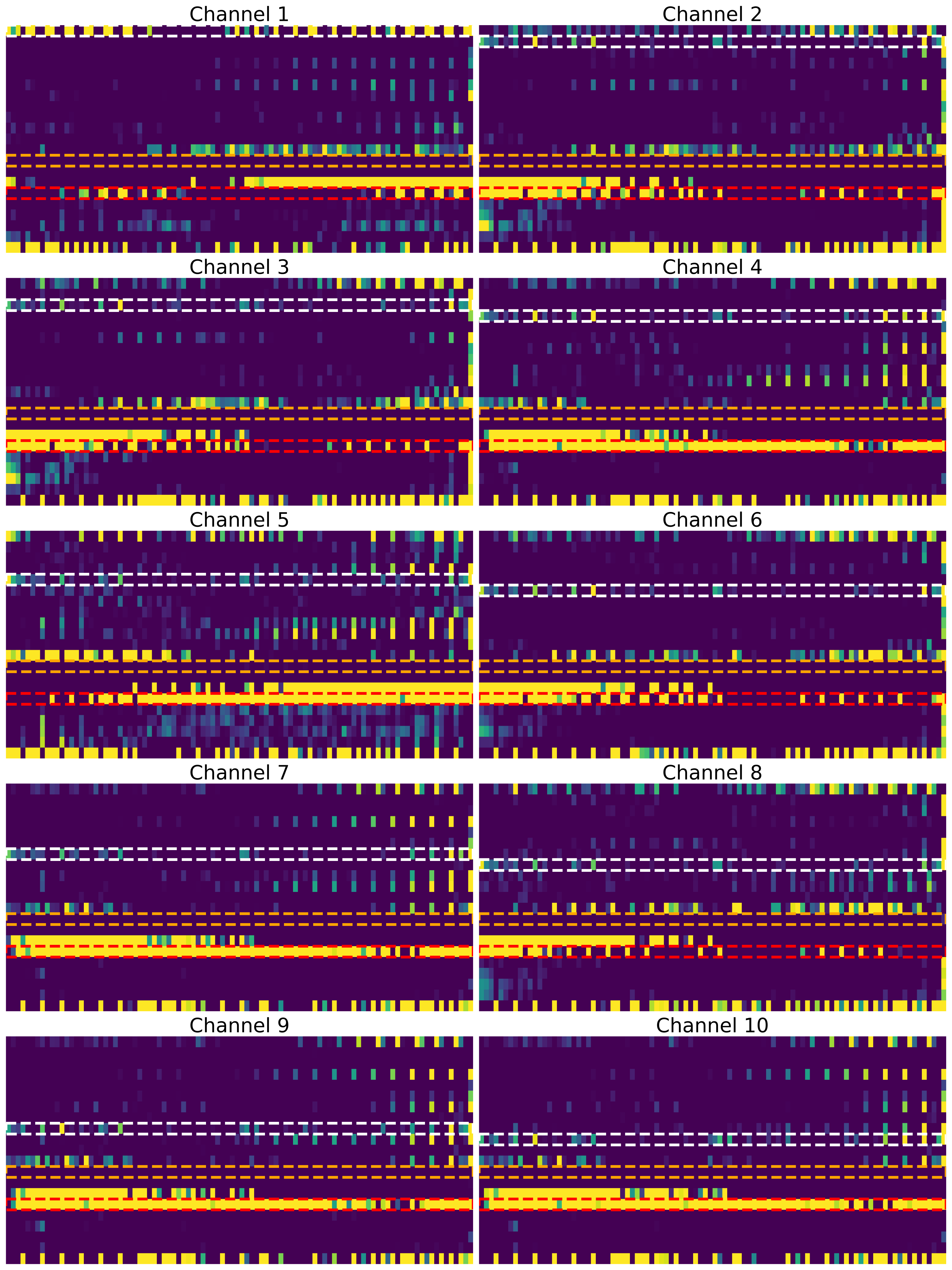}
    \caption{Visualization of individual receptive fields for the Weather dataset using ModernTCN.}
    \label{fig:Receptive_fields_viz}%
\end{figure}

In this section, we provide a detailed visualization of individual receptive fields to offer deeper insight into how convolutional networks learn inter-channel dependencies. We visualized the individual gradient values of ModernTCN on the Weather dataset, as shown in Figure \ref{fig:Receptive_fields_viz}. Following the methodology described in Section \ref{section:Individual_receptive_fields}, we sampled 100 data points and measured the aggregate influence by summing the absolute gradient values across each sample. The Weather dataset was selected for this analysis because its diverse meteorological variables provide a sufficiently complex environment to observe clear structural patterns. Each subplot in Figure \ref{fig:Receptive_fields_viz} corresponds to the gradient used to predict a specific output channel.

The visualization includes the following components: 
\begin{itemize}
  \item Self-Influence (White Box): This indicates the gradient values of an input sequence relative to its corresponding output sequence
  \item Pivot Channels (Red Box): We identified specific "pivot columns" that were highly and persistently activated across diverse output channels. This suggests that these variables contain critical information that drives the global state of the system.
  \item {Low Influencing Channels(Orange Box)}: This indicates the channels with persistently low gradient values across diverse output channels. 

\end{itemize}

This visualization provides empirical evidence the convolutional layers do not treat all variables equally; rather, they develop a preference for certain channels. This "implicit selection" mirrors the behavior of explicit attention mechanisms. To quantify these qualitative findings, we calculate the temporal variance of these gradients to derive the variance attention metric introduced in Section \ref{section:Individual_receptive_fields}.

\newpage
\subsection{Synthetic Datasets}
\label{appendix:synthetic}

This section provides the detailed experimental settings for the comparative analysis presented in Section 3.3. The parameters for the synthetic data generation in Equation 5 were configured as follows: the amplitude was set to $a \sim \text{Uniform}(-3, 3)$, the phase shift was set to $b \sim \text{Uniform}(0, 10)$, and the sequence length was fixed at $l = 200$.

To ensure a comparable architectural complexities between the evaluated models, we used the following configurations: (1) \textbf{Convolutional Model} Both the encoder and decoder consisted of two convolutional layers. Each layer increased the channel dimension by a factor of four by utilizing a kernel size of 10 and stride of 2. (2) \textbf{Linear Model} For linear configurations, matrix multiplications were designed to mimic the down- and up-sampling behavior of the convolutional layers. The encoder layers reduce the sequence length by half at each step, whereas the decoder layers project the hidden sequences back to their original length by doubling the sequence dimension.

All layers in both architectures were activated using the ReLU function. For the training phase, we generated a dataset of 10,000 samples, with an additional 1,000 samples reserved for evaluation. Each model was trained using the Adam optimizer with a consistent learning rate of 0.001.

\section{Parameter Sensitivity}
\begin{figure}[H]
    \centering
    \subfloat[\centering Hidden dimension]{{\includegraphics[width=.35\linewidth]{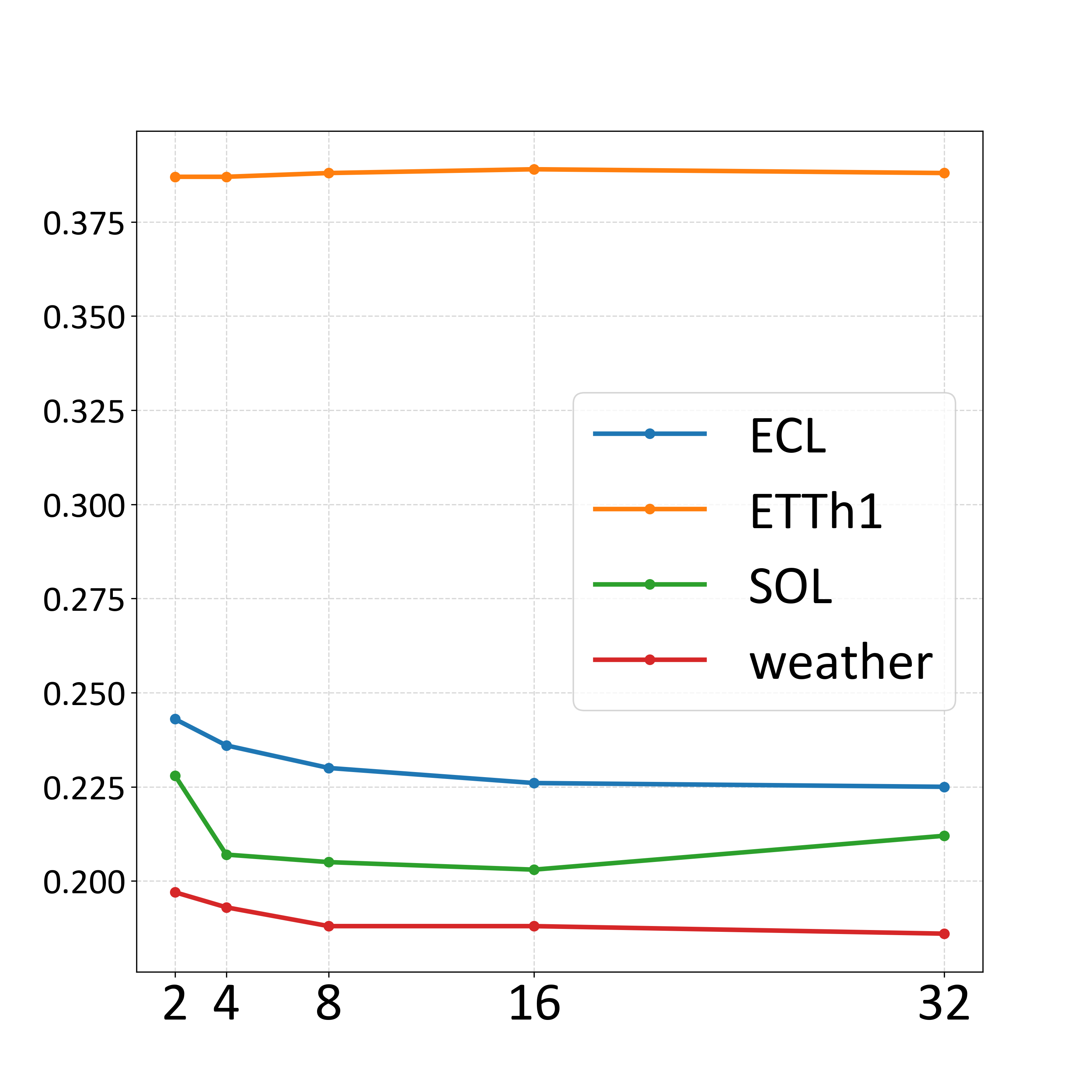} }}%
    \subfloat[\centering Number of layers]{{\includegraphics[width=.35\linewidth]{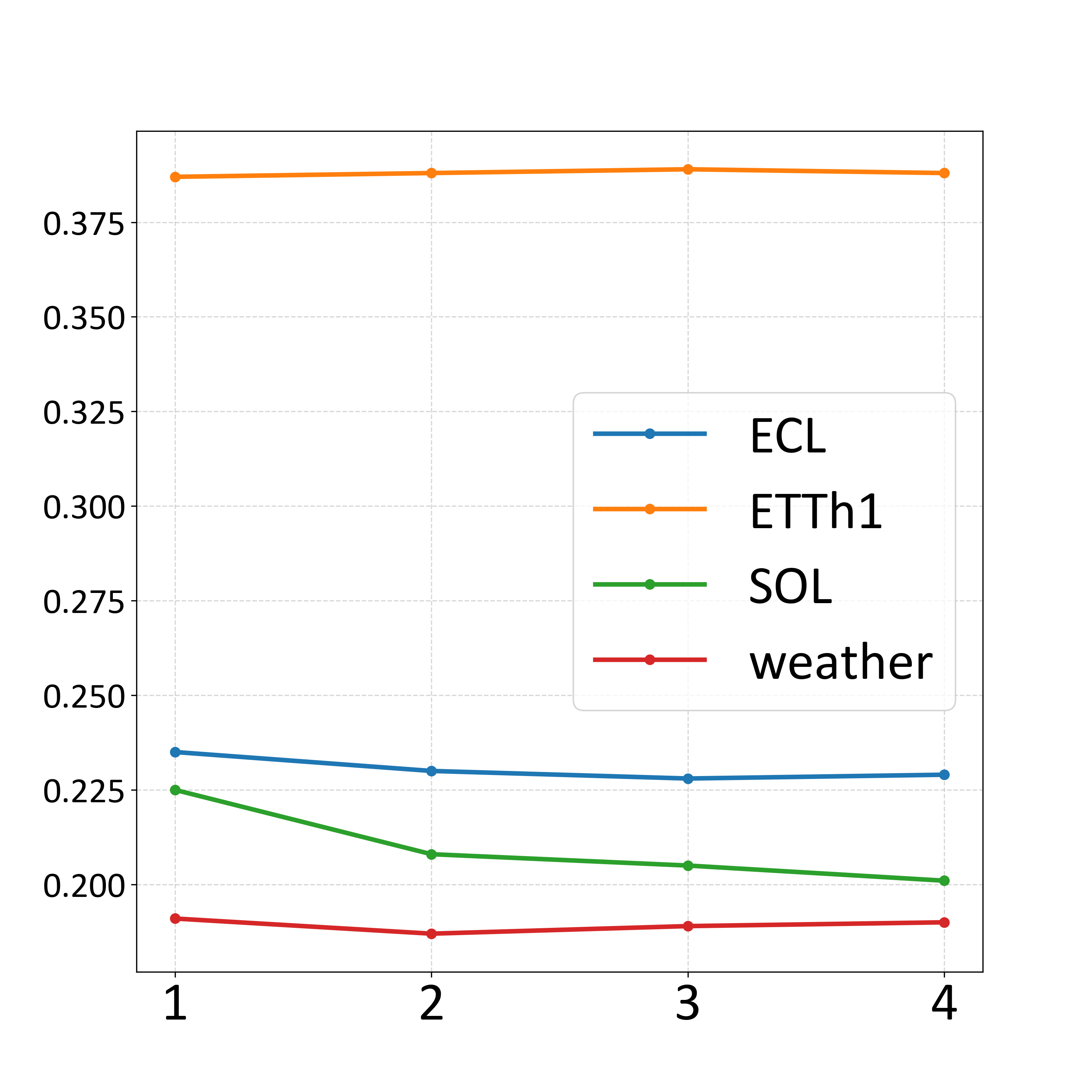} }}%
    \subfloat[\centering Input length]{{\includegraphics[width=.35\linewidth]{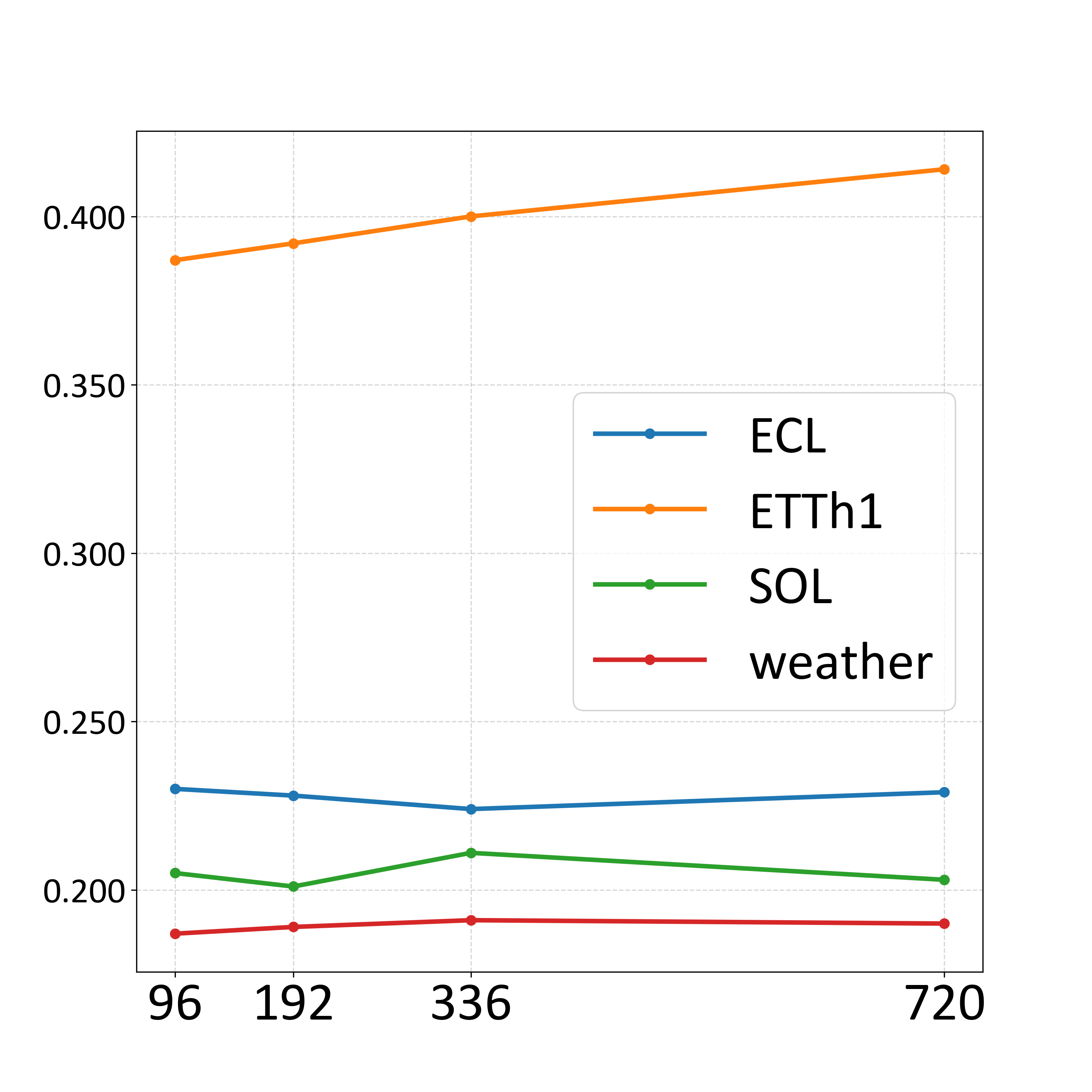} }}%
    \caption{Parameter sensitivity analysis across (a) hidden dimension size, (b) number of layers, and (c) look-back window length ($S$)}%
    \label{fig:parameter_sensitivity}%
\end{figure}

To evaluate the parameter sensitivity of ACFormer, we conducted experiments across the Electricity, ETTh1, Solar, and Weather datasets by varying three key hyperparameters: the number of kernels (multipliers of 2, 4, 8, 16, and 32), the number of encoder layers (1, 2, 3, and 4), and size of the look-back window (96, 192, 336, and 720). Except for the experiments that evaluated the look-back window, the sequence and prediction lengths were fixed at 96. As illustrated in Figure \ref{fig:parameter_sensitivity}, the results indicate that increasing the number of multipliers often leads to overfitting, which is similarly observed when the look-back window size is increased.

Unlike many existing models, ACFormer does not use linear embedding layers to project an input sequence onto a fixed dimension. Consequently, as the input length increased, the size of the hidden dimensions increased proportionally, which may have contributed to the observed sensitivity to window size. By contrast, the number of layers had a negligible effect on the predictive accuracy compared to the other two parameters, suggesting that the model's performance is driven more by its architectural dynamics than by depth. These findings highlight the importance of balancing the hidden dimension size with the input length to maintain an optimal performance without incurring redundant complexity.

\section{Full results}
\subsection{Long-term forecasting}
In this section, we provide the comprehensive benchmark results for multivariate long-term forecasting. Table \ref{tab:long_term_full} details the performance of ACFormer against all selected baselines across varying prediction lengths $P \in \{96, 192, 336, 720\}$. For each dataset and metric, the best-performing result is highlighted in bold, and the second-best result is underlined
\label{appendix:long_term_full}

\begin{table*}[h]
\centering
\small
\setlength{\tabcolsep}{2.2pt} 
\renewcommand{\arraystretch}{1.1}

\caption{Full results for the multivariate long-term forecasting task.}
\resizebox{\textwidth}{!}{
    \begin{tabular}{cc|cc|cc|cc|cc|cc|cc|cc|cc|cc}
    
    \toprule[1pt]
    \multicolumn{2}{c|}{\multirow{2}{*}{Models}} & 
    \multicolumn{2}{c|}{ACFormer} & 
    \multicolumn{2}{c|}{TimeMixer++} & 
    \multicolumn{2}{c|}{TimePro} & 
    \multicolumn{2}{c|}{Amplifier} & 
    \multicolumn{2}{c|}{iTransformer} & 
    \multicolumn{2}{c|}{ModernTCN} & 
    \multicolumn{2}{c|}{PatchTST} & 
    \multicolumn{2}{c|}{TimesNet} & 
    \multicolumn{2}{c|}{DLinear} \\
    & & 
    \multicolumn{2}{c|}{(Ours)} & \multicolumn{2}{c|}{(2025)} & \multicolumn{2}{c|}{(2025)} & \multicolumn{2}{c|}{(2025)} & \multicolumn{2}{c|}{(2024)} & \multicolumn{2}{c|}{(2024)} & \multicolumn{2}{c|}{(2023)} & \multicolumn{2}{c|}{(2023)} & \multicolumn{2}{c|}{(2023)}  \\
    
    \midrule
    \multicolumn{2}{c|}{Metric} & MSE & MAE & MSE & MAE & MSE & MAE & MSE & MAE & MSE & MAE & MSE & MAE & MSE & MAE & MSE & MAE & MSE & MAE\\
    
    \midrule
    \multirow{4}{*}{\rotatebox[origin=c]{90}{ecl}}
    &96&\textbf{0.139}&\underline{0.230}&0.164&0.252&\underline{0.141}&\textbf{0.229}&0.154&0.243&0.154&0.240&0.153&0.250&0.164&0.247&0.167&0.265&0.199&0.270\\
    &192&\textbf{0.157}&\textbf{0.246}&0.180&0.266&0.184&0.267&0.172&0.259&\underline{0.167}&\underline{0.252}&0.167&0.263&0.195&0.258&0.184&0.279&0.197&0.273\\
    &336&\textbf{0.172}&\underline{0.263}&0.193&0.277&\underline{0.173}&\textbf{0.261}&0.176&\underline{0.263}&0.182&0.268&0.176&0.273&0.190&0.273&0.213&0.306&0.209&0.288\\
    &720&\textbf{0.199}&\textbf{0.288}&0.224&0.300&0.224&0.300&0.210&0.293&0.211&0.293&\underline{0.200}&\underline{0.292}&0.226&0.303&0.241&0.322&0.245&0.320\\
    \midrule
    \multirow{4}{*}{\rotatebox[origin=c]{90}{etth1}}
    &96&\textbf{0.382}&\textbf{0.387}&0.397&0.398&0.387&0.402&\underline{0.385}&\textbf{0.387}&0.390&0.399&0.395&0.395&0.386&\textbf{0.387}&0.399&0.407&0.388&\underline{0.388}\\
    &192&\textbf{0.438}&\textbf{0.416}&0.451&0.429&0.444&0.431&0.442&0.421&\textbf{0.438}&0.429&0.462&0.436&\underline{0.439}&\underline{0.419}&0.455&0.440&0.446&0.424\\
    &336&\textbf{0.479}&\textbf{0.436}&0.494&0.447&\underline{0.484}&0.448&0.497&0.448&0.485&0.452&0.521&0.464&0.485&\underline{0.442}&0.516&0.479&0.491&0.449\\
    &720&\underline{0.488}&\textbf{0.459}&0.498&0.471&\textbf{0.480}&\underline{0.469}&0.529&0.480&0.498&0.478&0.553&0.508&0.507&0.477&0.534&0.495&0.508&0.485\\
    \midrule
    \multirow{4}{*}{\rotatebox[origin=c]{90}{etth2}}
    &96&\textbf{0.278}&\textbf{0.328}&0.295&0.338&0.290&0.333&0.285&\underline{0.332}&0.293&0.339&\underline{0.280}&0.335&0.287&0.334&0.315&0.355&0.293&0.344\\
    &192&\underline{0.362}&0.383&0.371&0.386&0.366&\underline{0.382}&\underline{0.362}&\underline{0.382}&0.377&0.392&0.373&0.387&\textbf{0.357}&\textbf{0.380}&0.383&0.397&0.377&0.396\\
    &336&\textbf{0.406}&\textbf{0.416}&0.422&0.426&0.409&\underline{0.417}&0.415&0.420&0.438&0.433&\underline{0.407}&0.418&0.437&0.428&0.447&0.434&0.449&0.451\\
    &720&\textbf{0.413}&\textbf{0.431}&0.430&0.441&\underline{0.419}&\underline{0.433}&0.427&0.439&0.438&0.447&\textbf{0.413}&0.435&0.439&0.443&0.422&0.436&0.610&0.548\\
    \midrule
    \multirow{4}{*}{\rotatebox[origin=c]{90}{ettm1}}
    &96&\textbf{0.307}&\textbf{0.337}&0.315&0.343&0.313&\underline{0.342}&0.316&\underline{0.342}&0.315&0.345&0.342&0.377&\underline{0.310}&\textbf{0.337}&0.394&0.389&0.332&0.351\\
    &192&\underline{0.362}&\underline{0.365}&0.363&0.369&0.376&0.376&0.369&0.367&0.366&0.370&0.373&0.396&\textbf{0.360}&\textbf{0.364}&0.452&0.426&0.376&0.374\\
    &336&\underline{0.388}&\textbf{0.385}&\textbf{0.386}&0.391&0.405&0.395&0.401&0.390&0.405&0.395&0.420&0.422&0.394&\underline{0.388}&0.471&0.439&0.406&0.395\\
    &720&\textbf{0.455}&\textbf{0.422}&\underline{0.459}&0.432&0.474&0.436&0.472&0.430&0.471&0.433&0.463&0.445&0.460&\underline{0.427}&0.538&0.476&0.469&0.432\\
    \midrule
    \multirow{4}{*}{\rotatebox[origin=c]{90}{ettm2}}
    &96&\textbf{0.169}&\textbf{0.247}&0.175&0.251&0.180&0.258&\underline{0.171}&\underline{0.249}&0.174&0.251&\textbf{0.169}&0.250&0.171&\underline{0.249}&0.187&0.261&0.183&0.257\\
    &192&\textbf{0.234}&\textbf{0.289}&0.245&0.298&0.245&0.300&\textbf{0.234}&\underline{0.292}&0.243&0.298&\underline{0.240}&0.299&0.235&0.293&0.260&0.307&0.245&0.302\\
    &336&\underline{0.295}&\textbf{0.329}&0.296&0.332&0.302&0.335&0.300&0.334&0.305&0.337&0.296&0.337&\textbf{0.292}&\underline{0.330}&0.322&0.346&0.307&0.348\\
    &720&\underline{0.395}&\underline{0.390}&0.440&0.418&0.403&0.393&\textbf{0.391}&\textbf{0.389}&0.403&0.394&0.453&0.424&0.402&0.400&0.431&0.407&0.413&0.419\\
    \midrule
    \multirow{4}{*}{\rotatebox[origin=c]{90}{sol}}
    &96&0.189&\textbf{0.205}&\textbf{0.184}&\underline{0.207}&0.283&0.286&\underline{0.185}&0.208&0.200&0.209&0.206&0.233&0.219&0.230&0.230&0.227&0.286&0.294\\
    &192&\textbf{0.221}&\textbf{0.224}&\underline{0.231}&0.236&0.333&0.315&0.234&0.235&0.236&\underline{0.233}&0.241&0.252&0.249&0.248&0.271&0.252&0.318&0.313\\
    &336&\textbf{0.244}&\textbf{0.241}&0.261&0.255&0.362&0.312&\textbf{0.244}&\underline{0.243}&\underline{0.253}&0.248&0.264&0.265&0.274&0.263&0.305&0.281&0.363&0.327\\
    &720&\textbf{0.256}&\underline{0.248}&0.263&0.256&0.361&0.307&\underline{0.257}&0.249&\underline{0.257}&\textbf{0.246}&0.267&0.266&0.278&0.263&0.308&0.277&0.374&0.322\\
    \midrule
    \multirow{4}{*}{\rotatebox[origin=c]{90}{traffic}}
    &96&0.437&\underline{0.257}&0.461&0.263&\underline{0.410}&\textbf{0.253}&0.447&0.283&\textbf{0.408}&0.268&0.456&0.289&0.628&0.307&0.622&0.312&0.671&0.366\\
    &192&0.457&\underline{0.270}&0.474&0.272&\underline{0.432}&\textbf{0.263}&0.457&0.285&\textbf{0.430}&0.276&0.460&0.286&1.038&0.311&0.642&0.319&0.625&0.343\\
    &336&0.471&\underline{0.276}&0.498&0.279&\underline{0.449}&\textbf{0.270}&0.471&0.292&\textbf{0.446}&0.281&0.475&0.292&0.568&0.324&0.652&0.326&0.629&0.345\\
    &720&0.508&\textbf{0.295}&0.547&\underline{0.299}&0.507&0.309&\underline{0.506}&0.314&\textbf{0.481}&0.300&0.511&0.311&0.558&0.323&0.698&0.349&0.660&0.365\\
    \midrule
    \multirow{4}{*}{\rotatebox[origin=c]{90}{weather}}
    &96&\textbf{0.150}&\textbf{0.187}&0.162&0.197&0.202&0.237&0.153&\underline{0.190}&0.211&0.242&\underline{0.152}&0.194&0.161&0.197&0.162&0.204&0.208&0.232\\
    &192&\textbf{0.201}&\textbf{0.239}&0.211&0.243&0.239&0.265&0.205&\underline{0.240}&0.264&0.286&\underline{0.204}&0.242&0.207&0.240&0.215&0.252&0.245&0.267\\
    &336&\textbf{0.261}&\underline{0.284}&0.269&0.286&0.301&0.312&0.267&0.286&0.310&0.320&0.263&0.289&\underline{0.262}&\textbf{0.281}&0.295&0.306&0.288&0.307\\
    &720&\textbf{0.339}&\textbf{0.333}&0.352&0.341&0.369&0.354&0.344&0.338&0.377&0.362&0.344&0.342&\underline{0.343}&\underline{0.335}&0.386&0.369&0.347&0.358\\
    
    \bottomrule[1pt]
    \end{tabular}
}
\label{tab:long_term_full}
\end{table*}

\subsection{Short-term forecasting}
\label{appendix:short_term_full}
This section presents the detailed experimental results for the multivariate short-term forecasting task, as presented in Table \ref{tab:shor_term_full}. The performance was evaluated across several prediction horizons, specifically $P \in \{12, 24, 36, 48\}$ for the ETT datasets and $P \in \{12, 24, 48, 96\}$ for the PEMS subsets. Following the established convention, the best results are indicated in bold, and the second-best results are underlined.

\begin{table*}[h]
\centering
\small
\setlength{\tabcolsep}{2.2pt} 
\renewcommand{\arraystretch}{1.1}

\caption{Full results for the multivariate short-term forecasting task.}
\resizebox{\textwidth}{!}{
    \begin{tabular}{cc|cc|cc|cc|cc|cc|cc|cc|cc|cc}
    
    \toprule[1pt]
    \multicolumn{2}{c|}{\multirow{2}{*}{Models}} & 
    \multicolumn{2}{c|}{ACFormer} & 
    \multicolumn{2}{c|}{TimeMixer++} & 
    \multicolumn{2}{c|}{TimePro} & 
    \multicolumn{2}{c|}{Amplifier} & 
    \multicolumn{2}{c|}{iTransformer} & 
    \multicolumn{2}{c|}{ModernTCN} & 
    \multicolumn{2}{c|}{PatchTST} & 
    \multicolumn{2}{c|}{TimesNet} & 
    \multicolumn{2}{c|}{DLinear} \\
    & & 
    \multicolumn{2}{c|}{(Ours)} & \multicolumn{2}{c|}{(2025)} & \multicolumn{2}{c|}{(2025)} & \multicolumn{2}{c|}{(2025)} & \multicolumn{2}{c|}{(2024)} & \multicolumn{2}{c|}{(2024)} & \multicolumn{2}{c|}{(2023)} & \multicolumn{2}{c|}{(2023)} & \multicolumn{2}{c|}{(2023)}  \\
    
    \midrule
    \multicolumn{2}{c|}{Metric} & MSE & MAE & MSE & MAE & MSE & MAE & MSE & MAE & MSE & MAE & MSE & MAE & MSE & MAE & MSE & MAE & MSE & MAE\\

    \midrule
    \multirow{4}{*}{\rotatebox[origin=c]{90}{pems03}}
    &12&\textbf{0.063}&\textbf{0.165}&0.133&0.245&0.113&0.229&\underline{0.068}&\underline{0.173}&0.069&0.173&0.097&0.206&0.084&0.191&0.090&0.193&0.106&0.213\\
    &24&\textbf{0.084}&\textbf{0.190}&0.183&0.283&0.183&0.288&\underline{0.094}&\underline{0.202}&0.097&0.206&0.157&0.266&0.133&0.240&0.116&0.215&0.184&0.283\\
    &48&\textbf{0.133}&\textbf{0.239}&0.279&0.356&0.362&0.411&\underline{0.148}&\underline{0.256}&0.163&0.271&0.348&0.422&0.234&0.323&0.159&0.256&0.317&0.390\\
    &96&\textbf{0.187}&\textbf{0.293}&0.367&0.426&0.665&0.571&\underline{0.214}&\underline{0.314}&0.276&0.363&0.222&0.317&0.380&0.424&0.251&0.323&0.443&0.486\\
    \midrule
    \multirow{4}{*}{\rotatebox[origin=c]{90}{pems04}}
    &12&\textbf{0.073}&\textbf{0.174}&0.171&0.277&0.125&0.239&0.087&0.191&\underline{0.081}&\underline{0.186}&0.100&0.208&0.106&0.212&0.087&0.191&0.117&0.224\\
    &24&\textbf{0.090}&\textbf{0.196}&0.230&0.324&0.213&0.313&0.114&0.222&\underline{0.101}&0.210&0.149&0.259&0.172&0.273&0.103&\underline{0.209}&0.196&0.293\\
    &48&\textbf{0.123}&\textbf{0.233}&0.309&0.383&0.455&0.470&0.153&0.263&\underline{0.134}&\underline{0.245}&0.381&0.437&0.315&0.377&\underline{0.134}&\underline{0.245}&0.341&0.399\\
    &96&\textbf{0.171}&\underline{0.285}&0.362&0.433&0.800&0.644&0.202&0.309&\textbf{0.171}&\textbf{0.278}&\underline{0.189}&0.299&0.535&0.505&0.199&0.302&0.451&0.481\\
    \midrule
    \multirow{4}{*}{\rotatebox[origin=c]{90}{pems07}}
    &12&\textbf{0.057}&\textbf{0.149}&0.162&0.274&0.072&0.175&\underline{0.064}&\underline{0.161}&0.072&0.162&0.088&0.204&0.079&0.182&0.083&0.177&0.101&0.209\\
    &24&\textbf{0.076}&\textbf{0.171}&0.208&0.314&0.109&0.217&\underline{0.089}&0.190&0.093&\underline{0.189}&0.150&0.276&0.137&0.239&0.104&0.201&0.194&0.292\\
    &48&\textbf{0.116}&\textbf{0.211}&0.296&0.383&0.180&0.287&\underline{0.126}&0.229&0.126&\underline{0.224}&0.378&0.464&0.253&0.327&0.136&0.232&0.384&0.427\\
    &96&\underline{0.175}&\textbf{0.257}&0.385&0.453&0.326&0.401&0.178&0.278&\textbf{0.164}&\textbf{0.257}&0.208&0.296&0.414&0.428&0.182&\underline{0.266}&0.598&0.536\\
    \midrule
    \multirow{4}{*}{\rotatebox[origin=c]{90}{pems08}}
    &12&\textbf{0.071}&\textbf{0.168}&0.152&0.255&0.090&0.193&\underline{0.080}&\underline{0.182}&0.082&0.183&0.110&0.218&0.094&0.198&0.111&0.204&0.113&0.218\\
    &24&\textbf{0.095}&\textbf{0.193}&0.206&0.300&0.135&0.239&0.119&0.225&\underline{0.117}&\underline{0.217}&0.215&0.317&0.153&0.252&0.139&0.232&0.197&0.291\\
    &48&\textbf{0.147}&\textbf{0.243}&0.310&0.378&0.238&0.327&\underline{0.183}&0.279&0.209&0.292&0.456&0.477&0.281&0.344&0.192&\underline{0.275}&0.388&0.416\\
    &96&\textbf{0.251}&\textbf{0.313}&0.408&0.432&0.447&0.458&0.325&0.370&0.385&0.396&0.315&0.368&0.468&0.449&\underline{0.299}&\underline{0.327}&0.664&0.532\\
    \midrule
    \multirow{4}{*}{\rotatebox[origin=c]{90}{etth1}}
    &12&\textbf{0.265}&\textbf{0.319}&0.271&0.324&0.315&0.365&0.276&0.329&0.279&0.331&\underline{0.268}&\textbf{0.319}&0.272&\underline{0.324}&0.299&0.347&0.279&0.328\\
    &24&\textbf{0.295}&\textbf{0.338}&0.300&0.345&0.327&0.368&\underline{0.296}&\underline{0.339}&0.305&0.350&0.297&\underline{0.339}&0.298&0.341&0.326&0.364&0.302&0.341\\
    &36&\underline{0.321}&\textbf{0.353}&0.325&0.360&0.352&0.384&\textbf{0.320}&\textbf{0.353}&0.333&0.366&0.324&\underline{0.355}&0.324&0.356&0.348&0.371&0.325&\textbf{0.353}\\
    &48&\textbf{0.336}&\textbf{0.361}&0.343&0.371&0.351&0.379&\textbf{0.336}&\textbf{0.361}&0.351&0.375&0.342&0.366&\underline{0.338}&0.364&0.394&0.395&\underline{0.338}&\underline{0.362}\\
    \midrule
    \multirow{4}{*}{\rotatebox[origin=c]{90}{etth2}}
    &12&0.130&0.226&0.131&\underline{0.225}&0.145&0.245&0.133&0.230&0.133&0.230&\textbf{0.127}&\textbf{0.224}&\underline{0.128}&\underline{0.225}&0.148&0.244&0.133&0.230\\
    &24&\underline{0.166}&0.252&0.174&0.256&0.170&\textbf{0.250}&0.169&0.254&0.175&0.259&\textbf{0.165}&0.252&\textbf{0.165}&\underline{0.251}&0.194&0.275&0.170&0.254\\
    &36&\underline{0.199}&0.276&0.203&0.276&0.201&\underline{0.273}&0.202&0.276&0.211&0.283&0.213&0.294&\textbf{0.197}&\textbf{0.272}&0.217&0.291&0.202&0.276\\
    &48&0.227&0.292&0.229&0.294&0.234&0.291&0.226&0.291&0.241&0.304&\textbf{0.218}&\underline{0.289}&\underline{0.219}&\textbf{0.287}&0.242&0.306&0.226&0.292\\
    
    \bottomrule[1pt]
    \end{tabular}
}
\label{tab:shor_term_full}
\end{table*}

\section{Explainability of convolution kernels}

The Independent Patch Expansion layer is specifically designed to reconstruct fine-grained temporal details by modeling the local dependencies unique to each variable. To investigate the efficacy of this approach, we visualized the learned convolution kernel weights and their corresponding input signals from the Weather dataset. We selected variables that represented distinct temporal behaviors: strong periodicity ($wv$, blue), weak periodicity ($sh$, orange), and sporadic, high-frequency fluctuations ($rain$, green).

\begin{figure}[H]
    \centering
    \subfloat[\centering Kernel weight values of $wv(m/s)$]{{\includegraphics[width=.3\linewidth]{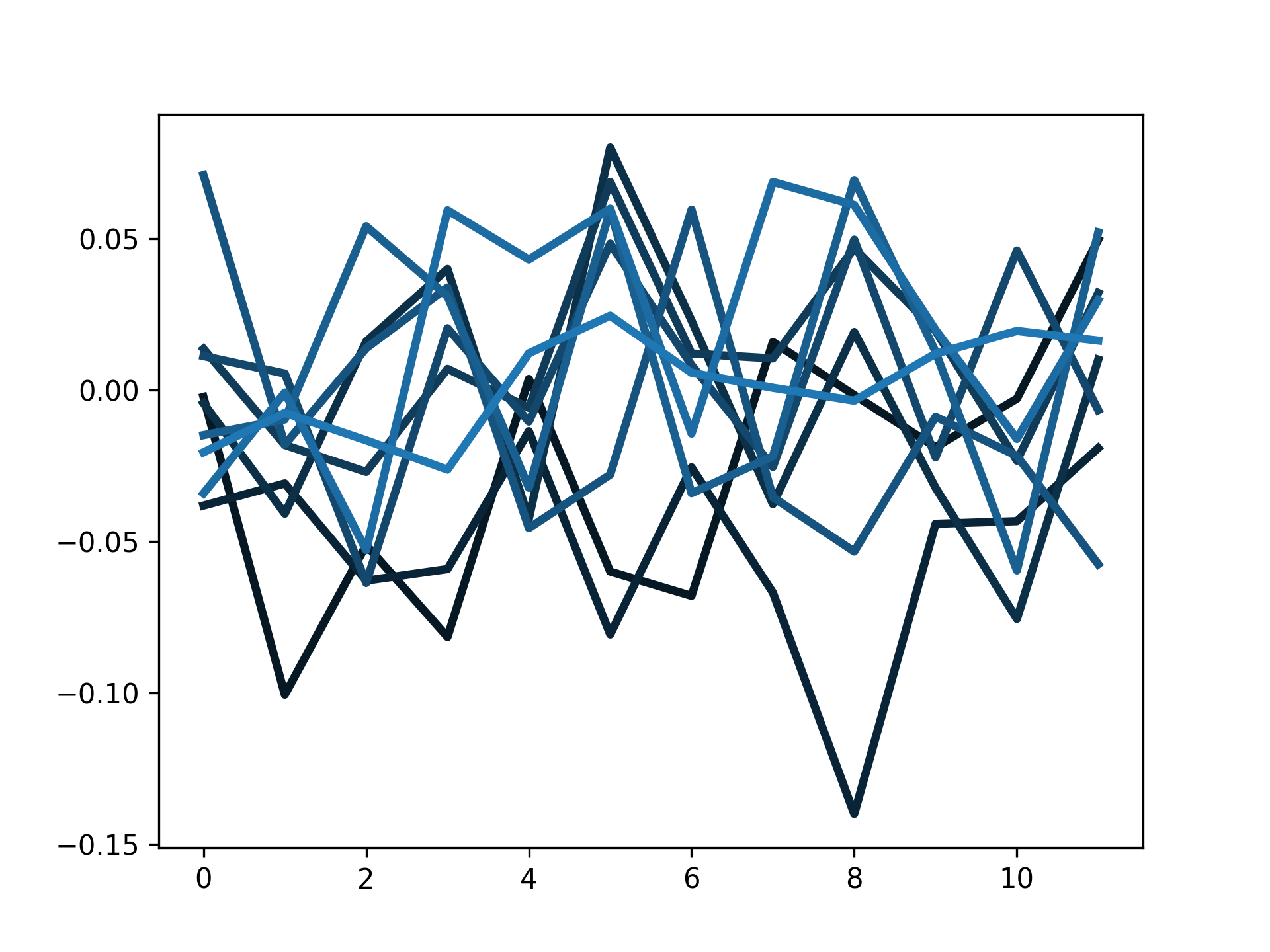} }}%
    \subfloat[\centering Kernel weight values of $sh(g/kg)$]{{\includegraphics[width=.3\linewidth]{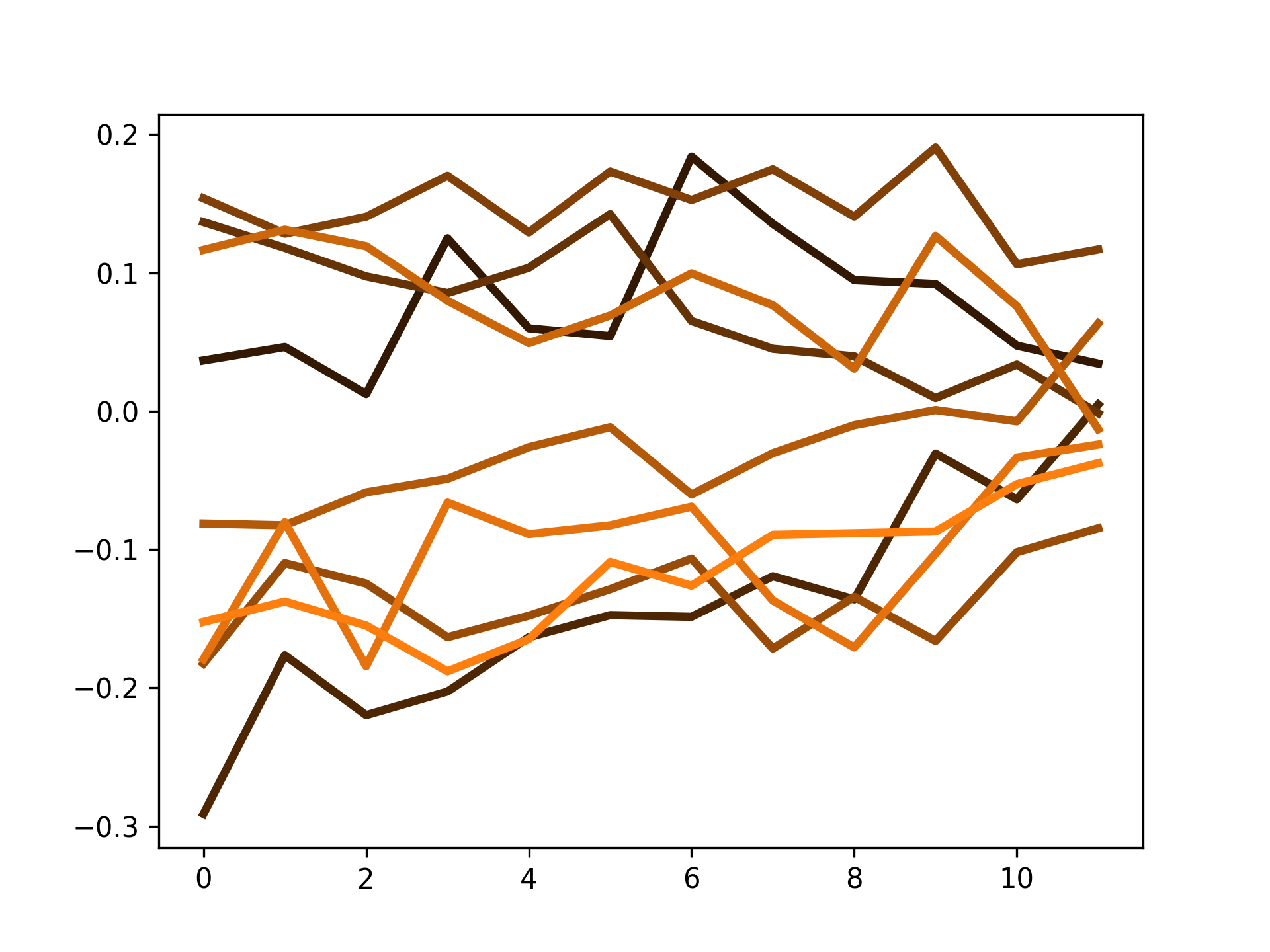} }}%
    \subfloat[\centering Kernel weight values of $rain(mm)$]{{\includegraphics[width=.3\linewidth]{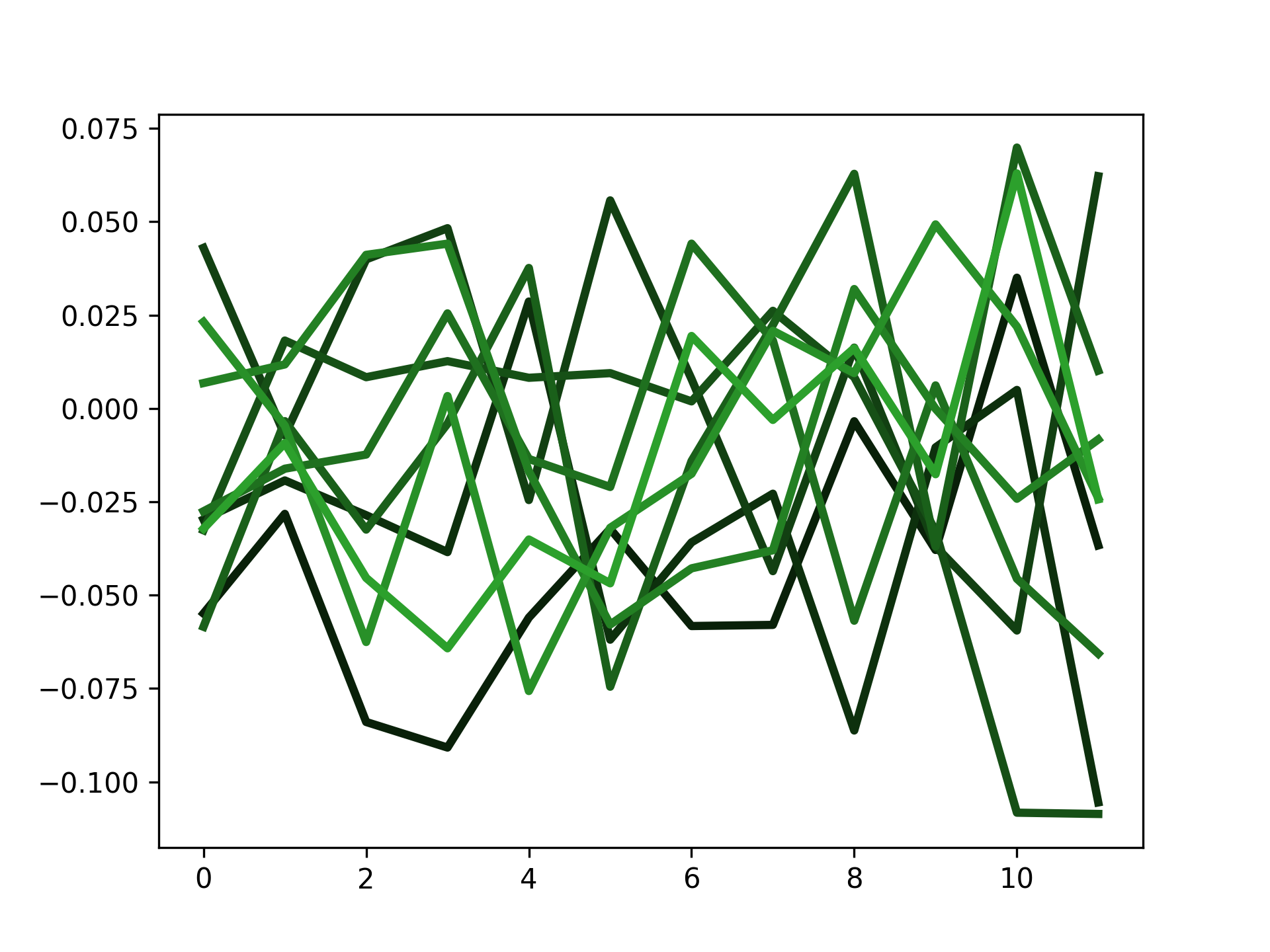} }}%
    \caption{Visualization of convolution kernels}%
    \label{fig:weather_kernel_weight}%
\end{figure}

As shown in Figure \ref{fig:weather_kernel_weight}, the independent kernels effectively adapt to the intrinsic characteristics of their respective variables. For variables with clear cyclical patterns, such as wind velocity ($wv$) and specific humidity ($sh$), the learned kernels exhibit a consistent, smooth shape that follows the underlying trend. This indicates that the expansion layer learns to amplify the stable oscillations that are essential for reconstructing seasonal patterns. By contrast, the kernels for sporadic variables such as rainfall ($rain$) exhibit irregular, non-uniform shapes across channels. 

This demonstrates that the model successfully identifies and amplifies non-linear, high-frequency nuances unique to specific channels ans details that would be smoothed out or lost by a linear projection. By allowing each channel to maintain unique expansion parameters, ACFormer performs a targeted high-frequency amplification, thereby enabling the recovery of non-linear nuances that are typically lost in standard models. This architecture effectively bridges the gap between universal pattern extraction and variable-specific detail reconstruction.

\begin{figure}[H]
    \centering
    \includegraphics[width=0.8\linewidth]{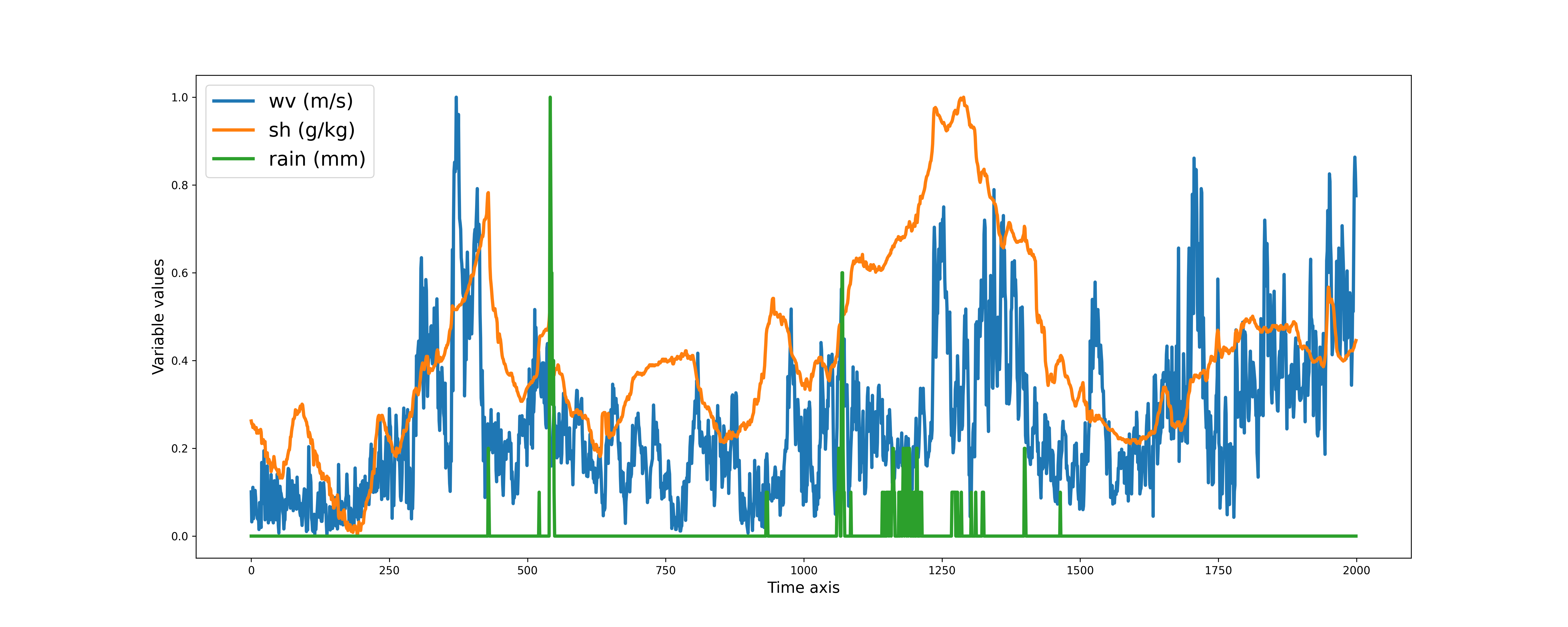}
    \caption{Visualization of the Weather dataset}
    \label{fig:weather_dataset_visualization}
\end{figure}


\end{document}